\documentclass[journal]{IEEEtran}

\newcommand{\our}{InstructMixup}

\usepackage{graphicx}
\usepackage{xcolor}
\usepackage{amsmath,amsfonts,amssymb}
\usepackage{booktabs}
\usepackage{multirow}
\usepackage{pifont}
\usepackage[ruled,vlined]{algorithm2e}
\usepackage{capt-of}

\usepackage{xcolor}

\providecommand{\our}{Ours}

\newcommand{\stddev}[1]{%
  _{\smash{\raisebox{0.15ex}{%
    \hbox{\textcolor{blue}{\tiny $\pm\,#1$}}}}}%
}

\newcommand{\result}[2]{%
  \ensuremath{#1\stddev{#2}}%
}

\newcommand{\bestresult}[2]{%
  \ensuremath{\mathbf{#1}\stddev{#2}}%
}

\newcommand{\secondresult}[2]{%
  \ensuremath{\underline{#1}\stddev{#2}}%
}

\newcommand{\estresult}[2]{%
  \ensuremath{#1^{\dagger}\stddev{#2}}%
}

\usepackage{graphicx}
\usepackage{booktabs}
\usepackage{xcolor}

\newcommand{\bestestresult}[2]{%
  \ensuremath{%
    \mathbf{#1}^{\dagger}_{%
      \smash{\textcolor{blue}{\scriptscriptstyle(\pm\,#2)}}%
    }%
  }%
}

\definecolor{darkgreen}{rgb}{0.0, 0.5, 0.0}
\definecolor{darkred}{rgb}{0.75, 0.0, 0.0}

\usepackage[colorlinks=true,citecolor=blue,linkcolor=blue,urlcolor=blue]{hyperref}
\usepackage{orcidlink}

\newcommand{\gain}[1]{#1}

\title{InstructMixup: Instruction-Guided Salient Patch Editing for Robust Data Augmentation}

\author{
Khawar Islam,
Arif Mahmood,
Xin Jin,
and Naveed Akhtar%
\thanks{Khawar Islam is with The University of Melbourne, Melbourne, Australia.}
\thanks{Arif Mahmood is with the Information Technology University, Lahore, Pakistan.}
\thanks{Xin Jin is with Westlake University, China.}
\thanks{Naveed Akhtar is with The University of Melbourne, Melbourne, Australia.}
\thanks{Corresponding author: Khawar Islam.}
}

\begin{document}

\maketitle

\begin{abstract}
In image and video technologies, data augmentation is widely used to improve the generalization of deep visual models, and mixup-based strategies that interpolate between samples have become the dominant approach. However, computing informative mixing regions adds substantial overhead, and blending content across different images frequently disrupts the semantic integrity of the resulting sample. We propose \our{}, a data augmentation method that constructs challenging yet label-consistent training samples entirely within a single visual sample. \our{} first extracts multi-scale salient patches from the sample using a lightweight saliency detector, refines each patch with an instruction-guided generative model, and blends the edited patch back into the non-salient regions of the same sample; because the generative edits are computed once and cached offline, this step adds negligible training cost. To further diversify the learned representation, \our{} injects self-similar fractal structure into the same salient regions at an adaptive ratio, so each training sample carries both fractal and non-fractal structure. We derive a second-order approximation of the resulting vicinal risk, showing that the method simultaneously enforces invariance to the generative edit and suppresses curvature along the perturbed salient directions, and we verify both predictions empirically. We evaluate on small to large backbones for instance Convolutional Neural Networks (CNNs), Vision Transformers (ViTs) and Vision-Language Foundational Models (VLMs) across seven benchmarks covering coarse- and fine-grained classification, robustness to corruption and occlusion, calibration, and transfer and self-supervised learning, InstructMixup outperforms nine competing augmentation methods, surpassing the strongest baseline across all benchmarks.
\end{abstract}


\section{Introduction}
\label{sec:intro}

\IEEEPARstart{M}{odern} Deep Neural Networks (DNNs) have grown large enough in scale and capacity to memorize their training data \cite{10106029,zhang2018mixup, cao2024survey, carratino2022mixup}. This capacity drives strong empirical accuracy, yet it also makes overfitting worse and widens the gap between training and test performance. In image and video technologies, data augmentation has become a primary tool for closing that gap \cite{11400596, kang2023guidedmixup, kim2020puzzle, kim2021co, takahashi2020ricap, wang2022isda}, and it is now standard across a range of practical systems.  
%image classification \cite{qin2025sumix, chen2022transmix, pu2024learnable}, object detection \cite{zoph2020learning}, and segmentation \cite{ghiasi2021simple, jin2025mergemix}. 
By broadening the training distribution, data augmentation  improves generalization to unseen inputs, reduces model collapse \cite{kang2023guidedmixup, xiao2023token, wang2024enhance}, helps re-balance under-represented classes \cite{zhang2023longtailed}, and makes models more resilient to distribution shift \cite{pinto2022using, jin2024survey, zhou2023domain, yu2024sfda}. Beyond input-space mixing, augmentation has been performed by randomly cropping and patching images and frames~\cite{takahashi2020ricap}, by translating deep features along semantic directions \cite{wang2022isda, pu2024learnable}, by blending superpixels with local and global context \cite{dornaika2023lgcoamix}, and by replacing patches to isolate causal features in few-shot regimes \cite{xu2023patchmix}.
\par
\begin{figure*}[t]
    \centering
    \includegraphics[width=0.90\linewidth]{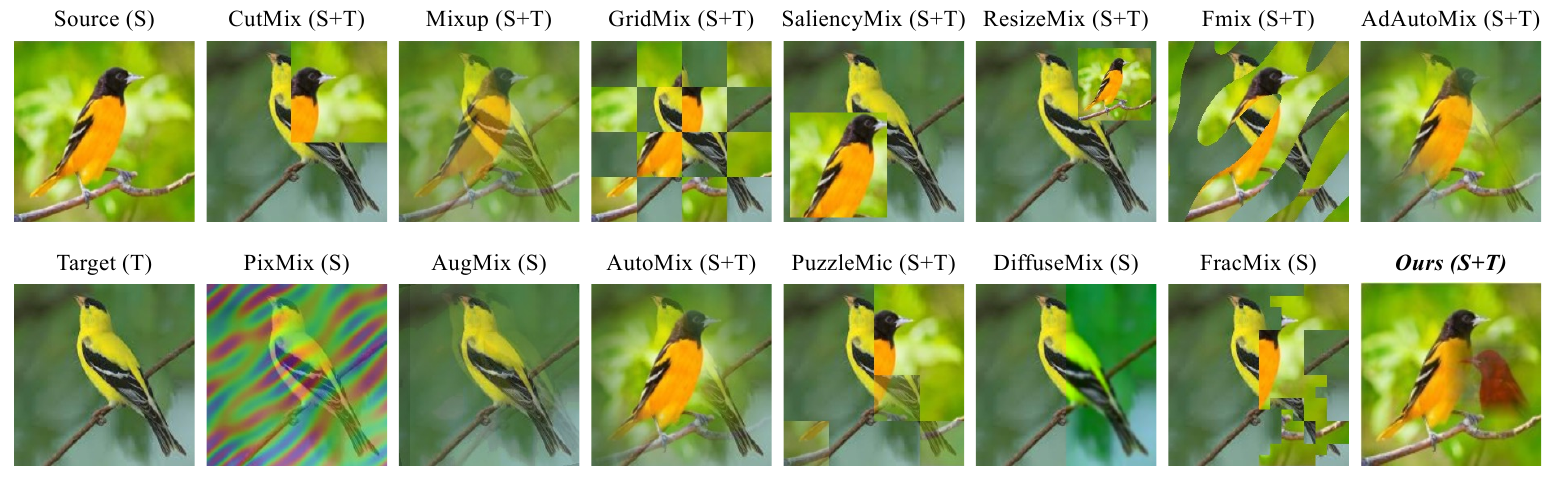}
    \vspace{-10pt}
    \caption{Representative augmentation samples created by different mixup methods. Each competing method constructs its sample from a source and a target frame, whereas \our{} operates entirely within a single frame.}
    \vspace{-10pt}
    \label{fig:2}
\end{figure*}

Any practical augmentation method must balance two demands: it should add diversity and robustness, yet keep the structural and semantic content of the data intact \cite{jin2024survey, huang2023ipmix, han2022cropmix, parast2025ddb, parast2025ghost, hendrycks2020augmix, verma2019manifold, jin2025mergemix}, all without driving up training cost \cite{kim2021co, kim2020puzzle}. We propose \our{}, an augmentation method that meets both demands, adding structural complexity and diversity to each sample while (see Figure~\ref{fig:2}).

Most task-agnostic \textit{mixup} methods synthesize new samples by combining random pairs of training visual samples \cite{zhang2018mixup}. CutMix~\cite{yun2019cutmix}, Manifold Mixup~\cite{verma2019manifold}, AlignMixup~\cite{venkataramanan2022alignmixup}, and ResizeMix~\cite{qin2020resizemix} each propose a different recipe for this pairing. Because the pairs are chosen at random, these methods can overwrite semantically important regions and break structural consistency. Saliency-guided variants such as SaliencyMix \cite{uddin2020saliencymix}, PuzzleMix \cite{kim2020puzzle}, Co-Mixup \cite{kim2021co}, and GuidedMixup \cite{kang2023guidedmixup} address this by steering the mix toward informative regions, but the saliency computation is expensive and pushes up both training time and hardware requirements \cite{kim2020puzzle, kim2021co}. AutoMix \cite{liu2022automix} and AdAutoMix \cite{qinadversarial} instead learn the mixing policy and label assignment automatically, yet they transfer poorly to Transformer backbones \cite{han2023vitsurvey} and remain compute-heavy. Fractal-based methods \cite{islam2024diffusemix, huang2023ipmix, hendrycks2022pixmix} take a different route but tend to overwrite real visual content and shift the data distribution. A further line of work uses generative models to edit or synthesize whole visual samples for added semantic variety~\cite{islam2024diffusemix, wang2024enhance}, though the methods are slow and can yield off-distribution or label-inconsistent results.
\par

\our{} is built to sidestep these issues, and it has two complementary parts: Self-Saliency ($S^2$) and \textit{AdaFrac}. The $S^2$ stage locates multi-scale salient patches in a sample and edits each one with an instruction-guided generative model, then mixes the edited patches back into the non-salient regions of the same sample. Keeping both the edit and the mix inside one frame  lets $S^2$ add targeted, label-consistent variation to the most discriminative regions without ever borrowing content from another sample, which in turn supports scale-invariant representation learning. The generative edits are produced offline, before training, so this step adds little runtime cost.

\textit{AdaFrac} builds on the salient regions that $S^2$ identifies. Where DiffuseMix \cite{islam2024diffusemix} spreads fractal texture over the whole image frame, \textit{AdaFrac} restricts self-similar fractal injection to those salient regions alone. This keeps the surrounding context clean while raising the structural complexity of the object itself, so every augmented sample ends up holding both fractal and non-fractal structure, gaining diversity without the distribution shift that global fractal mixing causes. \our{} also avoids committing to a single mixing rule. For each training instance it draws from several mixing modes at random, which widens the training distribution further and pushes the model toward more robust object representations. Evaluated on seven datasets against nine competing methods, \our{} delivers consistent improvements not only on clean test accuracy but also when the input is adversarially perturbed, partially occluded, or drawn from a downstream task distribution. We summarize our contributions as follows:

\begin{itemize}
\item \our{} introduces a multi-scale saliency-guided mixing stage that extracts informative patches from a sample and edits each one with an instruction-guided generative model before reinserting the edited patches into the non-salient regions of the same sample. As the edits are computed offline and cached in advance, this stage supplies controlled, label-consistent diversity to the most discriminative regions at negligible training cost.
\item We propose \textit{AdaFrac}, which injects self-similar fractal structure only within the salient regions identified by \our{}, raising the structural complexity of the object itself while leaving its surrounding context and data fidelity intact, avoiding the distribution shift caused by blending fractals over the whole sample.
\item We combine both components into \our{}, extend it with multi-mode mixing that draws at random from Mixup, CutMix, and ResizeMix, and derive a second-order approximation of the resulting vicinal risk that explains why the combination improves generalization.
\item We benchmark \our{} on seven datasets against nine state-of-the-art methods, reproducing 27 competing augmentation strategies for direct qualitative comparison. The evaluation spans coarse and fine-grained recognition, detection, transfer, self-supervised pre-training, calibration, and learning from limited data, where \our{} exceeds the strongest method.
\end{itemize}

This paper substantially extends our preliminary work, $S^{2}$-FracMix~\cite{islam2026s}. The conference version introduced a label-preserving self-saliency mixing method that relocate salient regions and injecting fractal perturbations within the same image. While effective, its augmentation capability is fundamentally limited to geometric transformations and structural perturbations of the existing visual content. Consequently, it cannot create new semantic appearance variations that naturally occur in real-world data, such as changes in texture, illumination, material, or style. To address this limitation, the journal version introduces instruction-guided generative patch editing, which synthesizes realistic, label-consistent appearance variations within the most discriminative regions while preserving the semantic identity of the object. The edited patches are generated offline, cached before training, and verified using a pretrained classifier to ensure label consistency, making the approach both computationally efficient and reliable. Beyond introducing a fundamentally new augmentation mechanism, we also develop an extended theoretical analysis by incorporating the generative editing process into the vicinal-risk formulation, providing a principled explanation for why the proposed framework improves generalization instead of only demonstrating empirical gains. Finally, unlike the conference version, which primarily evaluated image classification, the journal paper establishes that the proposed framework is architecture-independent and task-agnostic through extensive experiments on CNNs (ResNet-18, ResNeXt-50, ConvNeXt-T), Vision Transformers (Swin-T and ViT-B), and Vision-Language Models (CLIP), together with evaluations on classification, object detection, corruption and occlusion robustness, calibration, transfer learning, self-supervised learning, and few-shot learning.

% \begin{table}[t]
% \centering
% \caption{Qualitative comparison of augmentation properties across different methods.}
% \vspace{-7pt}
% \label{tab:novelty}

% \setlength{\tabcolsep}{3.2pt}
% \footnotesize
% \scalebox{0.90}{%
% \begin{tabular}{lcccccc}
% \toprule
% Method & Saliency & Self & Target & Multi-scale & Instruct. & Verified \\
% \midrule
% SaliencyMix~\cite{uddin2020saliencymix}
% & \cmark & \xmark & \xmark & \xmark & \xmark & \xmark \\

% PuzzleMix~\cite{kim2020puzzle}
% & \cmark & \xmark & \xmark & \xmark & \xmark & \xmark \\

% Co-Mixup~\cite{kim2021co}
% & \cmark & \xmark & \xmark & \xmark & \xmark & \xmark \\

% GuidedMixup~\cite{kang2023guidedmixup}
% & \cmark & \xmark & \xmark & \xmark & \xmark & \xmark \\

% SalfMix~\cite{choi2021salfmix}
% & \cmark & \cmark & \xmark & \xmark & \xmark & \xmark \\

% PixMix~\cite{hendrycks2022pixmix}
% & \xmark & \xmark & \xmark & \xmark & \xmark & \xmark \\

% IPMix~\cite{huang2023ipmix}
% & \xmark & \xmark & \xmark & \xmark & \xmark & \xmark \\

% DiffuseMix~\cite{islam2024diffusemix}
% & \xmark & \xmark & \xmark & \xmark & \xmark & \xmark \\

% $S^{2}$-FracMix~\cite{islam2026s}
% & \cmark & \cmark & \cmark & \cmark & \xmark & \xmark \\
% \midrule
% \textbf{\our{}}
% & \cmark & \cmark & \cmark & \cmark & \cmark & \cmark \\
% \bottomrule
% \end{tabular}}
% \end{table}

\section{Related Work}
\label{relatedWork}
% Researchers that developed data augmentation methods aims to improve the performance and generalization of machine learning models \cite{bishop2006pattern, lee2020network}. In the context of computer vision, simple transformations such as random cropping, horizontal flipping, color jittering, and rotation \cite{Krizhevsky2012ImageNet, han2022you} have been standard practice for decades, automatic color enhancement methods \cite{cubuk2018autoaugment, cubuk2020randaugment, hataya2020faster, li2020differentiable, suzuki2022teachaugment} to changed original ones. Early efforts in this domain demonstrated that small perturbations of input images could help models learn invariances and reduce overfitting, providing consistent gains across tasks such as image classification \cite{qinadversarial, goodfellow2014explaining}.

\subsection{Mixup Augmentation}
Mixup-based augmentation techniques are widely used to enrich training diversity through interpolation strategies \cite{lee2020smoothmix, yang2022recursivemix, hong2021stylemix}. Verma \textit{et al.} propose Manifold Mixup \cite{verma2019manifold}, which extends linear interpolation from the pixel space to hidden representations, encouraging smoother decision boundaries in the latent space. Yun \textit{et al.} introduce CutMix \cite{yun2019cutmix}, which pastes a rectangular crop from one image onto another so that the network learns to cope with missing evidence. FMix \cite{harris2020fmix} and GridMix \cite{baek2021gridmix} generalize the mask geometry beyond rectangles, while ResizeMix \cite{qin2020resizemix} shrinks an entire source image and overlays it onto the target, exposing the model to content at varying scales. SnapMix \cite{huang2021snapmix} targets fine-grained recognition, using class activation maps to weight the label of each mixed region by its semantic relevance. For Transformer backbones, TokenMix \cite{liu2022tokenmix} mixes at the token level and reweights labels using attention maps from a pretrained teacher, while MixPro \cite{zhao2023mixpro} combines patch-level masking with progressive attention labeling to better align mixed labels with the visible content ratio. More recently, Decoupled Mixup \cite{liu2024harnessing} formulates an efficient decoupled regularizer that mines discriminative features from hard mixed samples. Despite these advances, such methods enhance generalization primarily through inter-image mixing, without explicitly preserving semantic saliency, which can disrupt the discriminative regions most relevant to classification.

\subsection{Adversarial Mixup Methods}
Some influential lines of work integrate fractal images directly into the augmentation pipeline to improve model robustness. PixMix \cite{hendrycks2022pixmix} augments training images by blending them with synthetic images including fractals and feature visualizations. Building on this, IPMix \cite{huang2023ipmix} introduces multi-scale fractal mixing, where fractal patterns are inserted at pixel, patch, and image levels; the effectiveness of such patterns is grounded in their self-similar texture statistics \cite{vidivelli2023fractal}. Recent studies have further combined fractal augmentation with generative models: DiffuseMix \cite{islam2024diffusemix} blends a diffusion-generated image with the original, then overlays a fractal image, resulting in augmented views that are structurally complex. A complementary line of work instead perturbs images directly through adversarial optimization: AdvMask \cite{yang2023advmask} locates the image regions most influential to the classifier via a sparse adversarial attack module and occludes them to force the model to rely on additional discriminative cues. Despite their gains in robustness, these approaches apply fractal or adversarial perturbations globally, without adapting them to the semantic content of the image.

\subsection{Saliency-driven Mixup Augmentation}
A separate family of methods steers the mixing mask toward discriminative image content so that class evidence survives augmentation. SaliencyMix \cite{uddin2020saliencymix} and Attentive-CutMix \cite{walawalkar2020attentive} select the most informative crop of one image and paste it into another. PuzzleMix \cite{kim2020puzzle} formulates mixing as a transport problem that jointly respects saliency and local image statistics, and Co-Mixup \cite{kim2021co} generalizes this objective to batches of images under a supermodular diversity criterion. SAMix \cite{li2021boosting} splits the mixup objective into a locally emphasized and a globally constrained term for adaptive mixing, whereas SuperMix \cite{dabouei2021supermix} distills the mixing policy from a teacher network that identifies and merges salient evidence across images. Saliency Grafting \cite{park2021saliency} clips outlier attribution scores so that visually dominant but uninformative regions cannot dictate the mixed label, and GradSalMix \cite{hong2023gradsalmix} builds the mask directly from classifier gradients instead of an external saliency detector. LGCOAMix \cite{dornaika2023lgcoamix} exploits superpixel-level context at both local and global scales to keep object-part structure intact. Closest in spirit to our work, SalfMix \cite{choi2021salfmix} relocates the salient region of an image into its own background, and GuidedMixup \cite{kang2023guidedmixup} extracts critical local features via spectral-residual saliency. Even so, the majority of these methods still swap salient content \emph{between} images, and this cross-image transfer is precisely what breaks semantic consistency.

% \subsection{Automated Mixup Augmentation}
% Rather than hand-designing the mixing rule, another line of research learns it, closing the gap between augmentation heuristics and the training objective they are meant to serve. Smart Augmentation \cite{lemley2017smart} was an early attempt to co-train an augmentation network with the classifier, and Gontijo-Lopes \textit{et al.} \cite{gontijo2020affinity} later quantified what makes an augmentation useful through affinity and diversity measures, informing subsequent search-based methods. AutoMix \cite{liu2022automix} couples the generation of mixed samples with the classification loss in a single bi-level framework, and TransformMix \cite{cheung2024transformmix} learns the spatial transformation and mask jointly under supervision distilled from a pretrained teacher. Learned generators have also been used for this purpose, either adversarially \cite{zhao2020maximum} or with GANs \cite{antoniou2017data}, and AdAutoMix \cite{qinadversarial} synthesizes mixed samples adversarially to cope with domain shift. The price of this automation is a heavier training pipeline, and the learned policies frequently fail to carry over to new backbone architectures.

\begin{figure*}[t]
    \centering
    \includegraphics[width=0.90\linewidth]{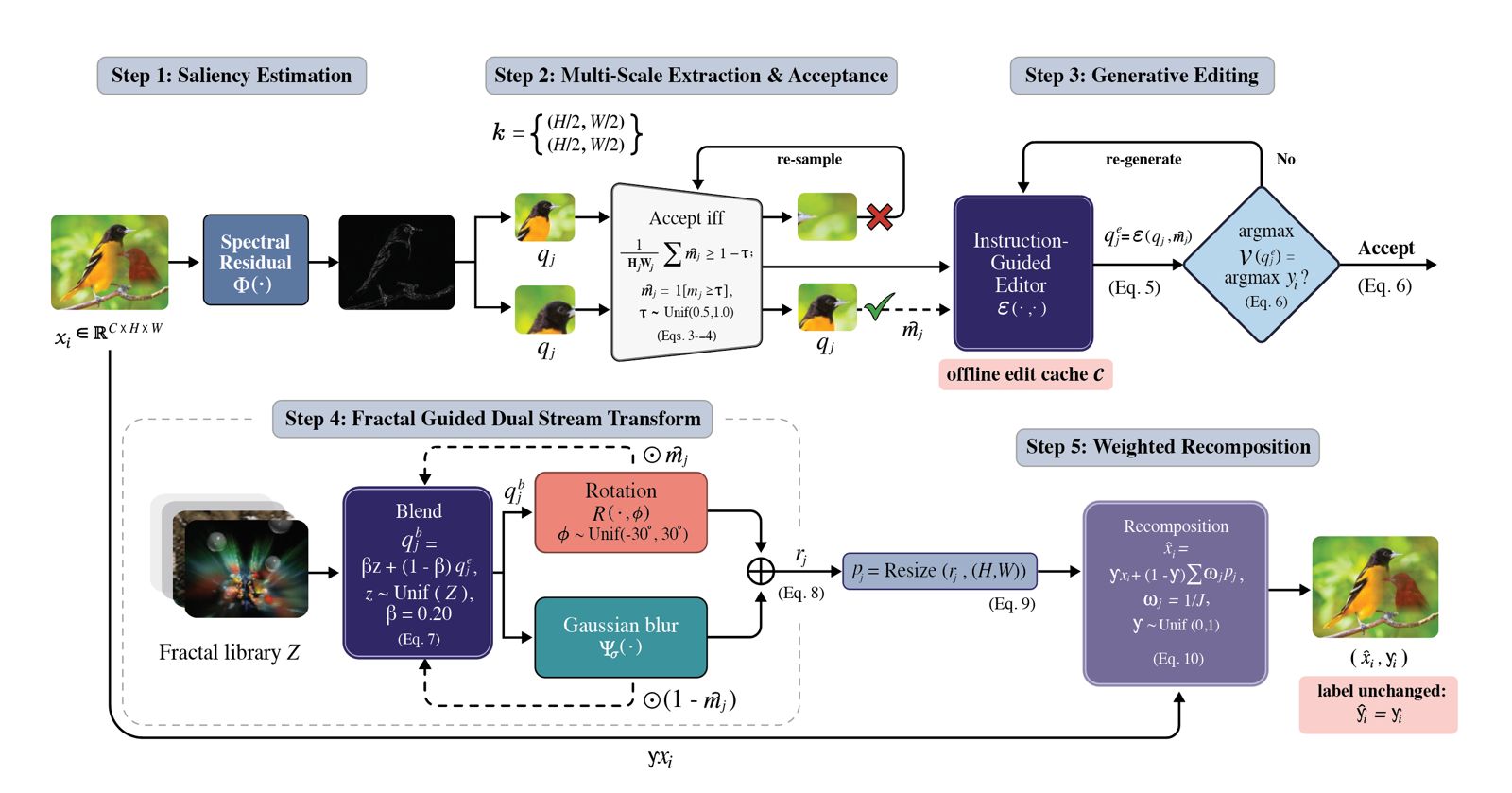}
    \vspace{-10pt}
\caption{\our{} overview. In the \our{} mode, the spectral-residual detector computes a saliency map and selects salient patches at two scales. Accepted patches are refined using an instruction-guided generative editor with label-consistency filtering. A fractal pattern is then blended with the patch, followed by rotation and Gaussian blur on salient and non-salient regions. Finally, the transformed patches are recomposed to produce the augmented image.}
    \vspace{-10pt}
    \label{fig:architecture}
\end{figure*}

\subsection{Generative Mixup Methods}
\label{sec:gen_mixup}
Recently, several generative mixup methods have also been proposed, such as DiffuseMix \cite{islam2024diffusemix}, DiffCoRe-Mix \cite{islam2025context} and DiffMix \cite{wang2024enhance}. These methods generate or edit images via contextual prompts to add meaningful features \cite{trabucco2023effective, wang2024improving}, and this idea has also been applied beyond classification, for instance to generate controllable synthetic training images for object detection \cite{fang2024data}. Although these methods have shown impressive gains, these techniques may produce out-of-distribution and unfaithful samples. It may become particularly important if a pre-trained generative model has not observed the training data distribution to be augmented \cite{zangdiffaug, parast2025ddb, parast2025ghost}. Another issue with generative mixup methods is their significant computational overhead due to generating new samples \cite{wang2024enhance, islam2024genmix}.
\par

In contrast to the generative mixup methods, we consider this work without requiring pre-trained models on the same or overlapping knowledge base. We also aim to design a probabilistic method with reduced computational overhead without affecting the generalization of the neural network. For this purpose, we propose \our{}, an approach that employs multi-scale and self-mixing method.

\section{Proposed Method}
\label{sec:PM}

\subsection{Overview}

\our{} encodes self-contained, multi-scale, saliency-guided augmentation within a single image frame while preserving object saliency and promoting structural diversity. Salient regions are located directly in the input via the spectral residual method~\cite{hou2007saliency}. These regions steer multi-scale patch extraction, so that augmentation strength is concentrated where it is least likely to erase class-discriminative content. Each extracted salient patch is then edited by an instruction-guided generative model, so that the most discriminative region of the sample  is \emph{refined} rather than merely copied verbatim -- introducing appearance diversity that spatial transformations alone cannot provide.

The edited patches are subsequently blended with a fractal texture and split into two complementary streams: the salient sub-region is rotated to encourage viewpoint invariance, while the surrounding non-salient sub-region is blurred to suppress background reliance. The resulting patches are resized and recomposed into the original sample, so that class-critical content is preserved while background or less-discriminative areas are perturbed to strengthen robust feature learning. As shown in Algorithm~\ref{alg:multiscale_mixup} and illustrated in Fig.~\ref{fig:architecture}, this pipeline is applied independently per frame, and -- unlike Mixup~\cite{guo2019mixup} and CutMix~\cite{yun2019cutmix} -- the original hard label is retained because the augmentation never imports content from a different class.

\subsection{Self-Saliency-Guided Multi-Scale Mixing}
\label{sec:self-saliency_method}

\textit{1) Problem Setup and Saliency Estimation:} Let $\mathcal{X} = \{(\mathbf{x}_i, \mathbf{y}_i)\}_{i=1}^{N}$ denote the training set, where each $\mathbf{x}_i \in \mathbb{R}^{C \times H \times W}$ is a $C$-channel image of spatial size $H \times W$ paired with a one-hot label $\mathbf{y}_i$. \our{} produces an augmented image $\hat{\mathbf{x}}_i$ while leaving the label unchanged, i.e., $\hat{\mathbf{y}}_i = \mathbf{y}_i$. A saliency map highlighting the regions most critical to model prediction is first computed as
\begin{equation}
    \mathbf{M}_i = \Phi(\mathbf{x}_i), \qquad \mathbf{M}_i \in \mathbb{R}^{1\times H \times W},
    \label{eq:saliency}
\end{equation}
where $\Phi(\cdot)$ denotes the spectral-residual saliency detector~\cite{hou2007saliency}. $\mathbf{M}_i$ subsequently governs both where candidate patches are accepted and how each accepted patch is later split into salient and non-salient streams.

\textit{2) Multi-Scale Candidate Extraction and Saliency-Based Acceptance:} Candidate patches are drawn from $\mathbf{x}_i$ at two scales, $\mathcal{K} = \{(H/2, W/2), (H/4, W/4)\}$. For a target scale $(H_j, W_j) \in \mathcal{K}$, the top-left corner $(u_j, v_j)$ is sampled uniformly,
\begin{equation}
    u_j \sim \text{Unif}(0, W-W_j), \qquad v_j \sim \text{Unif}(0, H-H_j),
    \label{eq:sample}
\end{equation}
and the candidate patch and its corresponding saliency sub-map are cropped as $\mathbf{q}_j = \mathbf{x}_i[u_j\!:\!u_j\!+\!W_j,\, v_j\!:\!v_j\!+\!H_j]$ and $\mathbf{m}_j = \mathbf{M}_i[u_j\!:\!u_j\!+\!W_j,\, v_j\!:\!v_j\!+\!H_j]$, respectively. We min--max normalize $\mathbf{m}_j$ to $[0,1]$ and draw a random acceptance threshold $\tau \sim \text{Unif}(0.5, 1.0)$ to obtain the binarized saliency mask
\begin{equation}
    \hat{\mathbf{m}}_j = \mathbb{1}\big[\mathbf{m}_j \ge \tau\big].
    \label{eq:binarize}
\end{equation}
The candidate patch $\mathbf{q}_j$ is accepted only if its salient-area ratio satisfies
\begin{equation}
    \frac{1}{H_j W_j}\sum \hat{\mathbf{m}}_j \;\ge\; (1-\tau);
    \label{eq:accept}
\end{equation}
otherwise it is discarded and re-sampled. Randomizing $\tau$ per patch, rather than fixing it, prevents the model from overfitting to a single saliency granularity and yields patches whose salient coverage varies across training iterations.

\textit{3) Saliency-Conditioned Generative Editing:} Naively reusing the raw salient crop, as in prior saliency-guided mixing methods, limits appearance diversity to whatever variation already exists in the source image itself. We instead let a generative model \emph{refine} the discriminative content, conditioned on the saliency mask so that class identity is preserved while low-level appearance (texture, lighting, style) is varied. Each accepted patch is passed through an instruction-guided generative editor $\mathcal{E}(\cdot,\cdot)$:
\begin{equation}
    \mathbf{q}_j^{e} = \mathcal{E}\big(\mathbf{q}_j,\, \hat{\mathbf{m}}_j\big),
    \label{eq:genedit}
\end{equation}
where $\hat{\mathbf{m}}_j$ conditions the edit so that $\mathcal{E}(\cdot,\cdot)$ concentrates its changes within the salient region while leaving the class-defining structure intact. To avoid incurring generative inference cost during training, every edit is produced \emph{offline} prior to training and stored in a cache $\mathcal{C}$, so that no generative model is invoked inside the training loop; at training time, $\mathbf{q}_j^{e}$ is simply retrieved from $\mathcal{C}$ given $(\mathbf{q}_j, \hat{\mathbf{m}}_j)$. We additionally discard any cached edit whose predicted label under a frozen, pretrained verifier $\mathcal{V}(\cdot)$ disagrees with $\mathbf{y}_i$,
\begin{equation}
    \mathbf{q}_j^{e} \;\text{is retained iff}\;\; \arg\max\, \mathcal{V}(\mathbf{q}_j^{e}) = \arg\max\, \mathbf{y}_i,
    \label{eq:consistency}
\end{equation}
which empirically preserves label consistency and removes edits that drift outside the source class (Sec.~\ref{sec:Ablation}).

\textit{4) Fractal-Guided Dual-Stream Transformation:} The edited patch $\mathbf{q}_j^{e}$ is blended with a texture $\mathbf{z}$ sampled from a fixed fractal library $\mathcal{Z}$ (constructed once, prior to training; see Sec.~\ref{sub:fractal_minmodes}),
\begin{equation}
    \mathbf{q}_j^{b} = \beta\, \mathbf{z} + (1-\beta)\, \mathbf{q}_j^{e}, \qquad \mathbf{z} \sim \text{Unif}(\mathcal{Z}),
    \label{eq:fracmix}
\end{equation}
where $\beta \in [0,1]$ controls the fractal-blend strength; we set $\beta = 0.20$ by default (Sec.~\ref{sub:fractal_minmodes}). The fractal-enriched patch is then split into a salient stream and a non-salient stream, each undergoing a different spatial transformation:
\begin{equation}
    \mathbf{r}_j = \mathcal{R}(\mathbf{q}_j^{b}, \phi) \odot \hat{\mathbf{m}}_j \;+\; \Psi_\sigma(\mathbf{q}_j^{b}) \odot (1 - \hat{\mathbf{m}}_j),
    \label{eq:transform}
\end{equation}
where $\mathcal{R}(\cdot, \phi)$ rotates its input by an angle $\phi \sim \text{Unif}(-30^{\circ}, 30^{\circ})$, $\Psi_\sigma(\cdot)$ applies Gaussian blurring with kernel bandwidth $\sigma$, and $\odot$ denotes the element-wise (Hadamard) product. Rotating the salient stream encourages viewpoint-invariant recognition of the discriminative content, while blurring the non-salient stream discourages the network from relying on background texture. The transformed patch is then resized back to the original image resolution,
\begin{equation}
    \mathbf{p}_j = \text{Resize}\big(\mathbf{r}_j, (H, W)\big),
    \label{eq:resize}
\end{equation}
where $\text{Resize}(\cdot,(H,W))$ applies bilinear interpolation so that $\mathbf{p}_j$ can be directly recomposed with the full-resolution input in the next step.

\textit{5) Weighted Recomposition:} Let $J$ denote the number of patches accepted for image $\mathbf{x}_i$ (across both scales in $\mathcal{K}$). The resized patches $\{\mathbf{p}_j\}_{j=1}^{J}$ are mixed back into the original image via a weighted sum,
\begin{equation}
    \hat{\mathbf{x}}_i = \gamma\, \mathbf{x}_i + (1-\gamma)\sum_{j=1}^{J} \omega_j\, \mathbf{p}_j,
    \label{eq:mix}
\end{equation}
where $\omega_j = 1/J$ so that $\sum_{j=1}^{J}\omega_j = 1$, and $\gamma \sim \text{Unif}(0,1)$ is a global blending coefficient resampled independently for every image. Because $\hat{\mathbf{y}}_i = \mathbf{y}_i$ throughout, \our{} trains the model to be robust to a combination of spatial transformations and generative appearance changes without the label-mixing bookkeeping, or the combinatorial mask optimization, required by SaliencyMix~\cite{uddin2020saliencymix}, PuzzleMix~\cite{kim2020puzzle}, and Co-Mixup~\cite{kim2021co}. The resulting procedure remains computationally lightweight at training time -- the only added cost relative to a standard augmentation pipeline is a saliency-map computation and a cache lookup -- while still synthesizing a diverse set of mixing outcomes that preserve the semantic content needed for the classifier to learn the correct label.

\begin{algorithm}[t]
\caption{\our{} Augmentation}
\label{alg:multiscale_mixup}
\small
\SetAlgoLined
\DontPrintSemicolon

\textbf{Require:} mini-batch $\mathcal{X}_b = \{(\mathbf{x}_i, \mathbf{y}_i)\}_{i=1}^{b}$; scale set $\mathcal{K}$; fractal library $\mathcal{Z}$; blend factor $\beta$; edit cache $\mathcal{C}$\\
\textbf{Ensure:} augmented batch $\hat{\mathcal{X}}_b = \{(\hat{\mathbf{x}}_i, \mathbf{y}_i)\}_{i=1}^{b}$\\

\ForEach{$(\mathbf{x}_i, \mathbf{y}_i) \in \mathcal{X}_b$}{
    $\mathbf{M}_i \leftarrow \Phi(\mathbf{x}_i)$ \tcp{Step 1: saliency map, Eq.~\ref{eq:saliency}}
    $\mathbf{p}_{\text{sum}} \leftarrow \mathbf{0}_{H\times W}$,\quad $J \leftarrow 0$\;

    \ForEach{scale $(H_j, W_j) \in \mathcal{K}$}{
        $u_j \leftarrow \texttt{uniform}(0, W-W_j)$,\quad
        $v_j \leftarrow \texttt{uniform}(0, H-H_j)$\;

        $\mathbf{q}_j \leftarrow \mathbf{x}_i[u_j\!:\!u_j\!+\!W_j,\, v_j\!:\!v_j\!+\!H_j]$ \tcp{Step 2: candidate patch}
        $\mathbf{m}_j \leftarrow \mathbf{M}_i[u_j\!:\!u_j\!+\!W_j,\, v_j\!:\!v_j\!+\!H_j]$\;
        $\mathbf{m}_j \leftarrow (\mathbf{m}_j - \min \mathbf{m}_j)/(\max \mathbf{m}_j - \min \mathbf{m}_j)$\;

        $\tau \leftarrow \texttt{uniform}(0.50, 1.0)$,\quad
        $\hat{\mathbf{m}}_j \leftarrow \mathbb{1}[\mathbf{m}_j \ge \tau]$\;

        \If{$\texttt{sum}(\hat{\mathbf{m}}_j)/(H_j W_j) < (1-\tau)$}{
            \texttt{reject} $\mathbf{q}_j$ \tcp{fails Eq.~\ref{eq:accept}}
        }
        \Else{
            $\mathbf{q}_j^{e} \leftarrow \texttt{fetch}(\mathcal{C}, \mathbf{q}_j, \hat{\mathbf{m}}_j)$ \tcp{Step 3: cached generative edit, Eq.~\ref{eq:genedit}}
            $\mathbf{z} \leftarrow \texttt{uniform}(\mathcal{Z})$\;
            $\mathbf{q}_j^{b} \leftarrow \beta \mathbf{z} + (1-\beta)\mathbf{q}_j^{e}$ \tcp{Step 4a: fractal blend, Eq.~\ref{eq:fracmix}}
            $\phi \leftarrow \texttt{uniform}(-30, 30)$\;
            $\mathbf{r}_j \leftarrow \mathcal{R}(\mathbf{q}_j^{b},\phi)\odot\hat{\mathbf{m}}_j + \Psi_\sigma(\mathbf{q}_j^{b})\odot(1-\hat{\mathbf{m}}_j)$ \tcp{Step 4b: dual-stream transform, Eq.~\ref{eq:transform}}
            $\mathbf{p}_j \leftarrow \texttt{Resize}(\mathbf{r}_j, (H,W))$ \tcp{Step 4c: resize, Eq.~\ref{eq:resize}}
            $\mathbf{p}_{\text{sum}} \leftarrow \mathbf{p}_{\text{sum}} + \mathbf{p}_j$,\quad $J \leftarrow J + 1$\;
        }
    }

    $\gamma \leftarrow \texttt{uniform}(0,1)$\;
    $\hat{\mathbf{x}}_i \leftarrow \gamma \mathbf{x}_i + \dfrac{1-\gamma}{J}\,\mathbf{p}_{\text{sum}}$ \tcp{Step 5: recomposition, Eq.~\ref{eq:mix}}
    $\hat{\mathcal{X}}_b \leftarrow \hat{\mathcal{X}}_b \cup \{(\hat{\mathbf{x}}_i, \mathbf{y}_i)\}$\;
}

\textbf{return} $\hat{\mathcal{X}}_b$
\end{algorithm}

\subsection{Multi-Mode High-Level Mixing}
\label{sub:fractal_minmodes}

Conventional fractal-blending methods apply the fractal texture over the entire input image. \our{} instead confines fractal blending to the salient patches selected in Sec.~\ref{sec:self-saliency_method}, so that the fractal texture perturbs appearance without overwhelming class-discriminative structure. The fractal library $\mathcal{Z}$ is constructed once, prior to training, and reused across all images; for every accepted and generatively edited patch $\mathbf{q}_j^{e}$, a texture $\mathbf{z}$ is drawn uniformly from $\mathcal{Z}$ and blended according to Eq.~\ref{eq:fracmix}. We choose the blend strength $\beta$ by sweeping a handful of candidate values on withheld validation data (Sec.~\ref{main_paper:Hyperparameters}, Fig.~\ref{fig:hyparameter}) and fix $\beta = 0.20$ in all subsequent experiments; the fractal-enriched patch $\mathbf{q}_j^{b}$ then proceeds through the dual-stream transformation of Eq.~\ref{eq:transform}, the resizing of Eq.~\ref{eq:resize}, and the recomposition of Eq.~\ref{eq:mix}. A qualitative visualization of every stage of this process is deferred to Figure~\ref{fig:aug_vis} in Sec.~\ref{sec:Ablation}.

Existing mixing-based augmentations typically commit to a \emph{single} strategy throughout training -- e.g., Mixup~\cite{guo2019mixup}, CutMix~\cite{yun2019cutmix}, ResizeMix~\cite{qin2020resizemix}, PuzzleMix~\cite{kim2020puzzle}, and GuidedMixup~\cite{kang2023guidedmixup}. We find that restricting training to a single low-level mixing mode narrows the supervisory signal the network is exposed to and, in turn, weakens downstream performance (Sec.~\ref{sec:Ablation}). We therefore incorporate \emph{multiple} low-level mixing modes into the pipeline, combining Mixup~\cite{guo2019mixup}, CutMix~\cite{yun2019cutmix}, and ResizeMix~\cite{qin2020resizemix} jointly, at a high level, alongside \our{}. One mode is selected uniformly at random for each training instance, which exposes the model to a richer variety of mixed inputs and encourages complementary forms of regularization; we quantify this effect in Sec.~\ref{sec:Ablation}.

\subsection{Training Objective}
\label{sec:trainobj}

Unlike Mixup and CutMix, which interpolate both the image \emph{and} the label, \our{} interpolates only the image while preserving the original hard label, since class-discriminative content is refined and retained rather than blended in from a different class. The augmented pair $(\hat{\mathbf{x}}_i, \mathbf{y}_i)$ is therefore used directly with the standard cross-entropy loss,
\begin{equation}
    \mathcal{L} = \frac{1}{N}\sum_{i=1}^{N} \mathcal{L}_{\text{CE}}\big(h_\theta(\hat{\mathbf{x}}_i),\, \mathbf{y}_i\big),
    \label{eq:loss}
\end{equation}
where $h_\theta(\cdot)$ denotes the target classifier with parameters $\theta$, and $\mathcal{L}_{\text{CE}}(\cdot,\cdot)$ is the cross-entropy loss between the predicted class distribution and the ground-truth label. Because no label mixing occurs, \our{} can be dropped into any standard classification training loop without modifying the loss function or requiring soft-label bookkeeping, distinguishing it from prior mixing-based augmentation strategies.

\section{A Second-Order Analysis of \our{}}
\label{lab:proof_analysis}

We now characterize the objective that \our{} implicitly optimizes. Viewing the saliency-guided generative edit together with the rotation and blur (Sec.~\ref{sec:self-saliency_method}) as a structured transform, and the fractal injection (Sec.~\ref{sub:fractal_minmodes}) as an additive localized perturbation, a second-order expansion of the vicinal risk yields two interpretable terms: the first enforces invariance to the structured transform, while the second penalizes loss curvature in directions confined to salient regions. We provide the analysis below. %This characterization yields two qualitative predictions, which we verify against our experiments in Sec.~\ref{sec:predictions}.

\subsection{Setup and Notation}
\label{sec:proof_setup}

Consider a sample $(\mathbf{x},\mathbf{y})\sim\mathcal{D}$, with $\mathbf{x}\in[0,1]^{C\times H\times W}$ a $C$-channel image at resolution $H\times W$ (following Sec.~\ref{sec:self-saliency_method}) and $\mathbf{y}$ its label. Write $h_\theta$ for the predictor of Eq.~\ref{eq:loss} and $\ell(\cdot,\cdot)$ for the training loss; in our implementation $\ell$ is the cross-entropy $\mathcal{L}_{\text{CE}}$ of Eq.~\ref{eq:loss}. To keep notation compact, we set
\begin{equation}
g(\mathbf{s}):=\ell\big(h_\theta(\mathbf{s}),\mathbf{y}\big),
\label{eq:gdef}
\end{equation}
where $g$ denotes the loss viewed as a function of the input $\mathbf{s}$ while the label $\mathbf{y}$ is held fixed. We denote by $\xi\sim\Pi$ the random augmentation state, where $\xi$ collects every source of randomness in the \our{} pipeline: the patch location sampled at each scale in $\mathcal{K}$ (Eq.~\ref{eq:sample}), the acceptance threshold and resulting saliency mask (Eqs.~\ref{eq:binarize}--\ref{eq:accept}), the retrieved generative edit (Eq.~\ref{eq:genedit}), the rotation angle $\phi$ and fractal sample (Eqs.~\ref{eq:fracmix}--\ref{eq:transform}), and the global blend coefficient $\gamma$ (Eq.~\ref{eq:mix}). We define the augmentation operator $A_\xi(\cdot)$ and the augmented sample
\begin{equation}
\mathbf{x}' := A_\xi(\mathbf{x}).
\label{eq:xprime}
\end{equation}

\subsection{Decomposing the Augmentation Operator}
\label{sec:proof_decomp}

Our analysis rests on splitting the augmentation operator into two parts: a structured transformation and an additive perturbation that acts only locally,
\begin{equation}
A_\xi(\mathbf{x}) = T_\xi(\mathbf{x}) + \Delta_\xi(\mathbf{x}),
\label{eq:augdecomp}
\end{equation}
where $T_\xi(\mathbf{x})\in[0,1]^{C\times H\times W}$ collects the structured part and $\Delta_\xi(\mathbf{x})\in\mathbb{R}^{C\times H\times W}$ the additive part. For \our{}, the structured part is saliency-guided: it comprises the generative edit of each salient patch together with the selective rotation/blur of Eq.~\ref{eq:transform}, accumulated over all accepted patches in Algorithm~\ref{alg:multiscale_mixup}.

Recall the saliency map $\mathbf{M}(\mathbf{x}):=\Phi(\mathbf{x})$ from Eq.~\ref{eq:saliency}, and let $\mathbf{B}_\xi\in\{0,1\}^{1\times H\times W}$ denote the full-resolution binary mask obtained by resizing and compositing the accepted patch-level masks $\{\hat{\mathbf{m}}_j\}$ (Eq.~\ref{eq:accept}) onto the image canvas. Let $\mathcal{E}_\xi(\mathbf{x})\in[0,1]^{C\times H\times W}$ denote the full-resolution image obtained by replacing each accepted salient region with its generatively edited counterpart $\mathbf{q}_j^{e}$ from Eq.~\ref{eq:genedit}, resized and composited back onto the canvas, while all non-salient content is left unchanged; the retrieved edit is part of $\xi$. Let $\mathcal{R}(\cdot,\phi)$ and $\Psi_\sigma(\cdot)$ denote the rotation and Gaussian-blur operators of Eq.~\ref{eq:transform}, with $\phi$ sampled as part of $\xi$. With the same global blend coefficient $\gamma\sim\text{Uniform}(0,1)$ as in Eq.~\ref{eq:mix}, we define
\begin{equation}
\begin{aligned}
T_\xi(\mathbf{x})=\gamma \mathbf{x}+(1-\gamma)\Big(&\mathcal{R}\big(\mathcal{E}_\xi(\mathbf{x}),\phi\big)\odot \mathbf{B}_\xi\\
&+ \Psi_\sigma(\mathbf{x})\odot (1-\mathbf{B}_\xi)\Big),
\end{aligned}
\label{eq:Txi}
\end{equation}
where $\odot$ denotes the element-wise (Hadamard) product; Eq.~\ref{eq:Txi} is the full-image aggregate of the per-patch dual-stream transform in Eq.~\ref{eq:transform}. Because the generative edit is structured and not zero-mean, it is absorbed into $T_\xi$ rather than into the perturbation term; the model is thus driven to become invariant to it, exactly as for the rotation and blur.

The remaining term captures the fractal injection of Sec.~\ref{sub:fractal_minmodes}: textures enter only through the accepted regions, scaled by the same blend factor $\beta$ as in Eq.~\ref{eq:fracmix}. Let $\mathbf{Z}\in\mathbb{R}^{C\times H\times W}$ denote the full-resolution fractal pattern obtained by resizing and compositing the patch-level textures $\mathbf{z}$ sampled per Eq.~\ref{eq:fracmix} onto the canvas. Since natural fractal textures are not zero-mean, we write $\mathbf{Z}=\boldsymbol{\mu}_{\mathbf{Z}}+\tilde{\mathbf{Z}}$, where $\boldsymbol{\mu}_{\mathbf{Z}}=\mathbb{E}[\mathbf{Z}]$ and $\tilde{\mathbf{Z}}$ is the zero-mean residual with covariance $\boldsymbol{\Sigma}_{\mathbf{Z}}$, and we absorb the deterministic shift $(1-\gamma)\beta(\mathbf{B}_\xi\odot\boldsymbol{\mu}_{\mathbf{Z}})$ into $T_\xi$. The perturbation component is then
\begin{equation}
\Delta_\xi(\mathbf{x})=(1-\gamma)\beta\,\big(\mathbf{B}_\xi\odot \tilde{\mathbf{Z}}\big).
\label{eq:Deltaxi}
\end{equation}

\subsection{Second-Order Expansion of the Vicinal Risk}
\label{sec:proof_expansion}

Following the vicinal risk minimization (VRM) framework~\cite{chapelle2000vicinal}, the objective induced by training on augmented samples is
\begin{equation}
\begin{aligned}
\mathrm{VRM}(h_\theta)
&:=\mathbb{E}_{(\mathbf{x},\mathbf{y})\sim\mathcal{D}}\,\mathbb{E}_{\xi\sim\Pi}\big[\ell(h_\theta(A_\xi(\mathbf{x})),\mathbf{y})\big]\\
&=\mathbb{E}_{\mathbf{x},\mathbf{y}}\mathbb{E}_{\xi}\big[g\big(T_\xi(\mathbf{x})+\Delta_\xi(\mathbf{x})\big)\big].
\end{aligned}
\label{eq:vrmdef}
\end{equation}
For brevity, let
\begin{equation}
\mathbf{x}_T := T_\xi(\mathbf{x}), \qquad \boldsymbol{\delta} := \Delta_\xi(\mathbf{x}),
\label{eq:shorthand}
\end{equation}
so that $\mathbf{x}'=\mathbf{x}_T+\boldsymbol{\delta}$. By construction, $\tilde{\mathbf{Z}}$ is zero-mean and $\gamma$ is sampled independently of $(\mathbf{B}_\xi,\tilde{\mathbf{Z}})$, hence $\mathbb{E}_\xi[\boldsymbol{\delta}]=0$ without further assumptions. We additionally require that $T_\xi$ leave the semantic label unchanged; the generative edit $\mathcal{E}_\xi$ satisfies this by construction, since every cached edit must pass the offline verification of Eq.~\ref{eq:consistency} (Sec.~\ref{sec:self-saliency_method}). The sole role of this requirement is to license evaluating the loss against the original target $\mathbf{y}$ after transformation.

Expanding $g$ to second order around $\mathbf{x}_T$ and using $\mathbf{x}'=\mathbf{x}_T+\boldsymbol{\delta}$ gives
\begin{equation}
\begin{aligned}
g(\mathbf{x}_T+\boldsymbol{\delta})
\approx
g(\mathbf{x}_T)&+\nabla g(\mathbf{x}_T)^\top \boldsymbol{\delta}\\
&+\tfrac{1}{2}\boldsymbol{\delta}^\top H_g(\mathbf{x}_T)\boldsymbol{\delta},
\end{aligned}
\label{eq:taylor}
\end{equation}
where $\nabla g(\mathbf{x}_T)$ and $H_g(\mathbf{x}_T)$ are, respectively, the input-space gradient and Hessian of $g$ at $\mathbf{x}_T$. Taking expectation over $\xi$, the first-order term vanishes because the perturbation is zero-mean:
\begin{equation}
\mathbb{E}_\xi\big[\nabla g(\mathbf{x}_T)^\top \boldsymbol{\delta}\big]
=
\nabla g(\mathbf{x}_T)^\top \mathbb{E}_\xi[\boldsymbol{\delta}]
=0.
\label{eq:firstorder}
\end{equation}
Therefore, the expected loss under augmentation is approximately
\begin{equation}
\mathbb{E}_\xi[g(\mathbf{x}_T+\boldsymbol{\delta})]
\approx
\mathbb{E}_\xi[g(\mathbf{x}_T)]
+\tfrac{1}{2}\mathbb{E}_\xi\!\left[\boldsymbol{\delta}^\top H_g(\mathbf{x}_T)\boldsymbol{\delta}\right].
\label{eq:expectedtaylor}
\end{equation}
Taking expectation over $(\mathbf{x},\mathbf{y})\sim\mathcal{D}$ yields
\begin{equation}
\mathrm{VRM}(h_\theta)
\approx
\mathbb{E}_{\mathbf{x},\mathbf{y}}\mathbb{E}_\xi[g(\mathbf{x}_T)]
+\tfrac{1}{2}\mathbb{E}_{\mathbf{x},\mathbf{y}}\mathbb{E}_\xi\!\left[\boldsymbol{\delta}^\top H_g(\mathbf{x}_T)\boldsymbol{\delta}\right].
\label{eq:vrmtaylor}
\end{equation}
The second term becomes interpretable once the quadratic form is expressed as a trace via $\mathbf{v}^\top H\mathbf{v}=\mathrm{tr}(H\mathbf{v}\mathbf{v}^\top)$:
\begin{equation}
\mathbb{E}_\xi\!\left[\boldsymbol{\delta}^\top H_g(\mathbf{x}_T)\boldsymbol{\delta}\right]
=
\mathrm{tr}\!\left(H_g(\mathbf{x}_T)\,\mathbb{E}_\xi[\boldsymbol{\delta}\boldsymbol{\delta}^\top]\right).
\label{eq:tracerewrite}
\end{equation}
We now compute the second moment of the perturbation. Since $\boldsymbol{\delta}=(1-\gamma)\beta(\mathbf{B}_\xi\odot \tilde{\mathbf{Z}})$ and $\gamma\sim\text{Uniform}(0,1)$ is independent of $(\mathbf{B}_\xi,\tilde{\mathbf{Z}})$, we have $\mathbb{E}[(1-\gamma)^2]=\tfrac{1}{3}$, and thus
\begin{equation}
\mathbb{E}_\xi[\boldsymbol{\delta}\boldsymbol{\delta}^\top]
=
\frac{\beta^2}{3}\,\mathbb{E}_\xi\!\left[\mathbf{B}_\xi\,\boldsymbol{\Sigma}_{\mathbf{Z}}\,\mathbf{B}_\xi^\top\right].
\label{eq:deltacov}
\end{equation}
We define the mask-dependent covariance as the localized-perturbation covariance,
\begin{equation}
\boldsymbol{\Sigma}_{\mathrm{loc}}(\mathbf{x}):=\mathbb{E}_\xi\!\left[\mathbf{B}_\xi\,\boldsymbol{\Sigma}_{\mathbf{Z}}\,\mathbf{B}_\xi^\top\right].
\label{eq:sigmaloc}
\end{equation}
Substituting Eqs.~\ref{eq:deltacov} and~\ref{eq:sigmaloc} into Eq.~\ref{eq:tracerewrite} yields
\begin{equation}
\mathbb{E}_\xi\!\left[\boldsymbol{\delta}^\top H_g(\mathbf{x}_T)\boldsymbol{\delta}\right]
=
\frac{\beta^2}{3}\,\mathrm{tr}\!\left(H_g(\mathbf{x}_T)\,\boldsymbol{\Sigma}_{\mathrm{loc}}(\mathbf{x})\right).
\label{eq:quadfinal}
\end{equation}
Collecting terms, the vicinal risk is approximated to second order by
\begin{equation}
\begin{aligned}
\mathrm{VRM}(h_\theta)
\approx{}&
\underbrace{\mathbb{E}_{\mathbf{x},\mathbf{y}}\mathbb{E}_\xi\big[\ell(h_\theta(T_\xi(\mathbf{x})),\mathbf{y})\big]}_{\text{invariance term}}\\
&+
\underbrace{\frac{\beta^2}{6}\,\mathbb{E}_{\mathbf{x}}\!\left[\mathrm{tr}\!\left(H_g(T_\xi(\mathbf{x}))\,\boldsymbol{\Sigma}_{\mathrm{loc}}(\mathbf{x})\right)\right]}_{\text{saliency-local stability penalty}}.
\end{aligned}
\label{eq:vrmfinal}
\end{equation}

% \subsection{Interpretation and Empirical Predictions}
% \label{sec:predictions}

% Eq.~\ref{eq:vrmfinal} separates the effect of \our{} into two forces. The invariance term pushes the network to respond identically to an image and to its saliency-guided transform $T_\xi$, which now subsumes the generative patch edit alongside rotation and blur. The stability term is a Hessian penalty weighted by $\boldsymbol{\Sigma}_{\mathrm{loc}}(\mathbf{x})$, meaning that curvature is suppressed specifically along directions supported on the salient regions accepted in Algorithm~\ref{alg:multiscale_mixup}, rather than everywhere in the input space.

\begin{table*}[t]
\centering
\caption{Comparison of mixup-based augmentation methods across
coarse-grained and fine-grained image-classification benchmarks.
Panel A reports Top-1 performance (\%) on CIFAR-100, Tiny-ImageNet,
and ImageNet-1K. Panel B reports accuracy (\%) on Caltech Birds-200,
FGVC-Aircraft, and Stanford-Cars.}
\vspace{-7pt}
\label{tab:merged_mixup_benchmarks}

\setlength{\tabcolsep}{1.3pt}
\renewcommand{\arraystretch}{1.03}
\scriptsize

% ================================================================
% Panel A
% ================================================================
\scalebox{0.80}{%
\begin{tabular*}{1.25\textwidth}{
@{\extracolsep{\fill}}lcccccccccc@{}}
\toprule
\multicolumn{11}{l}{
\textit{A. Coarse-grained image-classification benchmarks}} \\
\midrule

\textbf{Method}
& \multicolumn{4}{c}{\textbf{CIFAR-100}}
& \multicolumn{2}{c}{\textbf{Tiny-ImageNet}}
& \multicolumn{4}{c}{\textbf{ImageNet-1K}} \\
\cmidrule(lr){2-5}
\cmidrule(lr){6-7}
\cmidrule(lr){8-11}

& ResNet-18
& ResNeXt-50
& Swin-T
& ConvNeXt-T
& ResNet-18
& ResNeXt-50
& ResNet-18
& ResNet-34
& ResNet-50
& ViT-B \\
\midrule

Vanilla
& \result{78.04}{0.14}
& \result{81.09}{0.10}
& \result{78.41}{0.13}
& \result{78.70}{0.12}
& \result{61.68}{0.19}
& \result{65.04}{0.15}
& \result{70.04}{0.08}
& \result{73.85}{0.07}
& \result{76.83}{0.06}
& \result{76.70}{0.09} \\

MixUp~\cite{zhang2018mixup}
& \result{79.12}{0.12}
& \result{82.10}{0.09}
& \result{76.78}{0.15}
& \result{81.13}{0.11}
& \result{63.86}{0.17}
& \result{66.36}{0.13}
& \result{69.98}{0.09}
& \result{73.97}{0.07}
& \result{77.12}{0.06}
& \result{80.80}{0.08} \\

CutMix~\cite{yun2019cutmix}
& \result{78.17}{0.13}
& \result{81.67}{0.10}
& \result{80.64}{0.12}
& \result{82.46}{0.10}
& \result{65.53}{0.16}
& \result{66.47}{0.12}
& \result{68.95}{0.10}
& \result{73.58}{0.08}
& \result{77.17}{0.06}
& \result{79.90}{0.09} \\

SaliencyMix~\cite{uddin2020saliencymix}
& \result{79.12}{0.11}
& \result{81.53}{0.09}
& \result{80.40}{0.11}
& \result{82.82}{0.09}
& \result{64.60}{0.15}
& \result{66.55}{0.11}
& \result{69.16}{0.09}
& \result{73.56}{0.07}
& \result{77.14}{0.05}
& -- \\

FMix~\cite{harris2020fmix}
& \result{79.69}{0.12}
& \result{81.90}{0.10}
& \result{80.72}{0.10}
& \result{81.79}{0.11}
& \result{63.47}{0.16}
& \result{65.08}{0.12}
& \result{69.96}{0.08}
& \result{74.08}{0.07}
& \result{77.19}{0.05}
& -- \\

PuzzleMix~\cite{kim2020puzzle}
& \result{81.13}{0.10}
& \result{82.85}{0.08}
& \result{80.33}{0.11}
& \result{82.29}{0.10}
& \result{65.81}{0.14}
& \result{67.83}{0.10}
& \result{70.12}{0.07}
& \result{74.26}{0.06}
& \result{77.54}{0.05}
& -- \\

ResizeMix~\cite{qin2020resizemix}
& \result{80.01}{0.11}
& \result{81.82}{0.09}
& \result{80.16}{0.12}
& \result{82.53}{0.10}
& \result{63.74}{0.15}
& \result{65.87}{0.11}
& \result{69.50}{0.08}
& \result{73.88}{0.06}
& \result{77.42}{0.05}
& -- \\

AutoMix~\cite{liu2022automix}
& \result{82.04}{0.09}
& \result{83.64}{0.07}
& \result{82.67}{0.09}
& \result{83.30}{0.08}
& \result{67.33}{0.12}
& \result{70.72}{0.09}
& \result{70.50}{0.07}
& \result{74.52}{0.05}
& \result{77.91}{0.04}
& -- \\

AdAutoMix~\cite{qinadversarial}
& \result{82.32}{0.08}
& \result{84.22}{0.06}
& \result{84.33}{0.08}
& \result{83.54}{0.07}
& \result{69.19}{0.11}
& \result{72.89}{0.08}
& \result{70.86}{0.06}
& \result{74.82}{0.05}
& \result{78.04}{0.04}
& -- \\

$S^{2}$-FracMix~\cite{islam2026s}
& \result{82.74}{0.07}
& \result{84.91}{0.06}
& \result{85.35}{0.07}
& \result{84.41}{0.06}
& \result{70.38}{0.10}
& \result{74.27}{0.07}
& \result{71.37}{0.05}
& \result{75.34}{0.04}
& \result{78.54}{0.04}
& \result{81.20}{0.06} \\

\midrule

\textbf{\our{}}
& \bestresult{82.94}{0.06}
& \bestresult{85.21}{0.05}
& \bestresult{85.55}{0.06}
& \bestresult{84.61}{0.05}
& \bestresult{70.68}{0.09}
& \bestresult{74.47}{0.06}
& \bestresult{71.57}{0.05}
& \bestresult{75.54}{0.04}
& \bestresult{78.74}{0.03}
& \bestresult{81.40}{0.05} \\

\bottomrule
\end{tabular*}%
}

\vspace{4pt}

% ================================================================
% Panel B
% ================================================================
\scalebox{0.80}{%
\begin{tabular*}{1.25\textwidth}{
@{\extracolsep{\fill}}lcccccccccc@{}}
%\toprule
\multicolumn{11}{l}{
\textit{B. Fine-grained visual-classification benchmarks}} \\
\midrule

\textbf{Method}
& \multicolumn{4}{c}{\textbf{Caltech Birds-200}}
& \multicolumn{2}{c}{\textbf{FGVC-Aircraft}}
& \multicolumn{4}{c}{\textbf{Stanford-Cars}} \\
\cmidrule(lr){2-5}
\cmidrule(lr){6-7}
\cmidrule(lr){8-11}

& ResNet-18
& ResNet-50
& Swin-T
& ConvNeXt-T
& ResNet-18
& ResNeXt-50
& ResNet-18
& ResNeXt-50
& ResNet-50
& ViT-B \\
\midrule

Vanilla
& \result{77.68}{0.18}
& \result{82.38}{0.14}
& \estresult{85.24}{0.12}
& \estresult{85.71}{0.11}
& \result{80.23}{0.17}
& \result{85.10}{0.13}
& \result{86.32}{0.15}
& \result{90.15}{0.11}
& \estresult{88.88}{0.10}
& \estresult{91.31}{0.09} \\

MixUp~\cite{zhang2018mixup}
& \result{78.39}{0.16}
& \result{82.98}{0.13}
& \estresult{85.66}{0.11}
& \estresult{86.08}{0.10}
& \result{79.52}{0.18}
& \result{85.18}{0.12}
& \result{86.27}{0.14}
& \result{90.81}{0.10}
& \estresult{89.45}{0.09}
& \estresult{91.36}{0.08} \\

CutMix~\cite{yun2019cutmix}
& \result{78.40}{0.17}
& \result{83.17}{0.12}
& \estresult{85.82}{0.11}
& \estresult{86.31}{0.10}
& \result{78.84}{0.19}
& \result{84.55}{0.13}
& \result{87.48}{0.13}
& \result{91.22}{0.10}
& \estresult{88.99}{0.10}
& \estresult{91.53}{0.08} \\

ManifoldMix~\cite{verma2019manifold}
& \result{79.76}{0.15}
& \result{83.76}{0.11}
& \estresult{86.35}{0.10}
& \estresult{86.79}{0.09}
& \result{80.68}{0.16}
& \result{86.60}{0.11}
& \result{85.88}{0.16}
& \result{90.20}{0.12}
& \estresult{89.20}{0.09}
& \estresult{91.72}{0.08} \\

SaliencyMix~\cite{uddin2020saliencymix}
& \result{77.95}{0.17}
& \result{81.71}{0.14}
& \estresult{84.67}{0.12}
& \estresult{85.13}{0.11}
& \result{80.02}{0.17}
& \result{84.31}{0.14}
& \result{86.48}{0.14}
& \result{90.60}{0.11}
& \estresult{89.13}{0.10}
& \estresult{91.56}{0.09} \\

FMix~\cite{harris2020fmix}
& \result{77.28}{0.18}
& \result{83.34}{0.12}
& \estresult{85.96}{0.11}
& \estresult{86.42}{0.10}
& \result{79.36}{0.18}
& \result{86.23}{0.11}
& \result{87.55}{0.13}
& \result{90.90}{0.10}
& \estresult{89.09}{0.09}
& \estresult{91.53}{0.08} \\

PuzzleMix~\cite{kim2020puzzle}
& \result{78.63}{0.14}
& \result{83.83}{0.11}
& \estresult{86.41}{0.10}
& \estresult{86.86}{0.09}
& \result{80.76}{0.15}
& \result{86.23}{0.10}
& \result{87.78}{0.12}
& \result{91.29}{0.09}
& \estresult{89.37}{0.08}
& \estresult{91.83}{0.07} \\

ResizeMix~\cite{qin2020resizemix}
& \result{78.50}{0.15}
& \result{83.41}{0.12}
& \estresult{86.02}{0.11}
& \estresult{86.48}{0.10}
& \result{78.10}{0.20}
& \result{84.08}{0.14}
& \result{88.17}{0.11}
& \result{91.36}{0.09}
& \estresult{89.21}{0.09}
& \estresult{91.64}{0.08} \\

AutoMix~\cite{liu2022automix}
& \result{79.87}{0.12}
& \result{83.88}{0.10}
& \estresult{86.74}{0.09}
& \estresult{87.16}{0.08}
& \result{81.37}{0.13}
& \result{86.72}{0.09}
& \result{88.89}{0.10}
& \result{91.38}{0.08}
& \estresult{88.71}{0.09}
& \estresult{92.51}{0.07} \\

AdAutoMix~\cite{qinadversarial}
& \result{80.88}{0.10}
& \result{84.57}{0.09}
& \estresult{87.42}{0.08}
& \estresult{87.85}{0.07}
& \result{81.73}{0.12}
& \result{87.16}{0.08}
& \result{89.19}{0.09}
& \result{91.59}{0.07}
& \estresult{89.65}{0.08}
& \estresult{91.38}{0.07} \\

$S^{2}$-FracMix~\cite{islam2026s}
& \result{81.84}{0.08}
& \result{85.73}{0.07}
& \estresult{88.56}{0.07}
& \estresult{88.98}{0.06}
& \result{82.81}{0.10}
& \result{88.34}{0.07}
& \result{90.56}{0.08}
& \result{92.86}{0.06}
& \estresult{90.85}{0.07}
& \estresult{92.86}{0.06} \\

\midrule

\textbf{\our{}}
& \bestresult{82.04}{0.07}
& \bestresult{85.93}{0.06}
& \bestestresult{88.76}{0.06}
& \bestestresult{89.18}{0.05}
& \bestresult{83.01}{0.09}
& \bestresult{88.54}{0.06}
& \bestresult{90.76}{0.07}
& \bestresult{93.06}{0.05}
& \bestestresult{91.05}{0.06}
& \bestestresult{93.06}{0.05} \\

\bottomrule
\end{tabular*}%
}

\vspace{-10pt}
\end{table*}
%\input{Tabs/fgvc}

% Two testable consequences follow, and our experiments confirm both. Because the stability penalty scales with $\beta^2$, increasing the fractal-blend strength should help up to a point and then hurt; Fig.~\ref{fig:hyparameter}(b) shows exactly this behavior, with a peak at $\beta=0.20$ and a decline by $\beta=0.50$. Moreover, the margin over AdAutoMix under corruption (\gain{2.70\%}) and the FGSM error reduction (\gain{3.44\%}) are both larger than the corresponding clean-accuracy margin (\gain{1.39\%}). Finally, local and global fractal injection simply instantiate different $\boldsymbol{\Sigma}_{\mathrm{loc}}(\mathbf{x})$; Table~\ref{tab:merged_ablation_efficiency} settles this comparison empirically in favor of the local variant.

\section{Experiments}
\label{sec:experiments}

\subsection{Datasets}
\label{sec:datasets}
Our evaluation covers seven benchmarks under the protocols established by prior
SOTA mixup methods~\cite{guo2019mixup, qinadversarial, yun2019cutmix, islam2024diffusemix}. For coarse-grained recognition we use CIFAR-100~\cite{Krizhevsky09learningmultiple} ($100$ classes of $32\times32$ images, $500$/$100$ per class for training/testing), Tiny-ImageNet~\cite{chrabaszcz2017downsampled} (a $200$-class ImageNet derivative at $64\times64$, with $500$/$50$/$50$ images per class for train/val/test), and the full ImageNet-1K~\cite{imagenet} ($\sim$$1.28$M training and $50{,}000$ validation images over $1{,}000$ categories). Fine-grained recognition is assessed on CUB-200-2011~\cite{WahCUB_200_2011} ($200$ bird species, $5{,}994$/$5{,}794$ train/test), FGVC-Aircraft~\cite{maji2013fine} ($100$ aircraft variants, $6{,}667$/$1{,}333$/$2{,}000$ train/val/test), and Stanford-Cars~\cite{KrauseStarkDengFei-Fei_3DRR2013} ($196$ car models, $8{,}144$/$8{,}041$ train/test). In addition, Oxford Flowers-102~\cite{nilsback2008automated} is used for the data-scarce and self-supervised settings, Pascal VOC~\cite{everingham2010pascal} for object detection, and the corrupted variants CIFAR-100-C~\cite{hendrycks2019robustness} and ImageNet-C~\cite{kim2021co} for robustness evaluation.

\subsection{Comparison Methods and Backbones}
\label{sec:comparison_methods}
We compare \our{} against nine competitive mixup
methods: Mixup~\cite{zhang2018mixup}, CutMix~\cite{yun2019cutmix},
ManifoldMix~\cite{verma2019manifold}, FMix~\cite{harris2020fmix},
ResizeMix~\cite{qin2020resizemix}, SaliencyMix~\cite{uddin2020saliencymix},
PuzzleMix~\cite{kim2020puzzle}, AutoMix~\cite{liu2022automix}, and
AdAutoMix~\cite{qinadversarial}, and additionally report computational overhead
against the timings in AdAutoMix~\cite{qinadversarial}. We also report the results of the preliminary version of our method, termed $S^{2}$-FracMix~\cite{islam2026s}, for reference.  To assess
generalizability, we span small- to large-scale backbones, including
\texttt{ResNet-18/34/50}~\cite{he2016deep} and
\texttt{ResNeXt-50}~\cite{xie2017aggregated}, the transformer architectures
\texttt{Swin Transformer}~\cite{liu2021swin} and \texttt{ConvNeXt}~\cite{liu2022convnet},
and the contrastive methods \texttt{MoCo~v2}~\cite{chen2020improved} and
\texttt{SimSiam}~\cite{chen2021exploring}. All experiments use the open-source
OpenMixup~\cite{li2022openmixup} and follow the evaluation protocol of
AdAutoMix~\cite{qinadversarial} for fair comparison.

%\input{Tabs/transfer}
% \begin{table}[t]
%   \centering
%       \caption{Top-1 accuracy and FGSM error of \texttt{ResNet-18} on CIFAR-100 and CIFAR-100-C.}
%       \vspace{-7pt}
%       \scalebox{0.90}{\input{Tabs/exp_robustness}}
%       \label{tab:4}
%       \vspace{-10pt}
% \end{table}

\subsection{Implementation Details}
\label{sec:implementation}

\begin{figure*}
    \centering
    \includegraphics[width=0.95\linewidth]{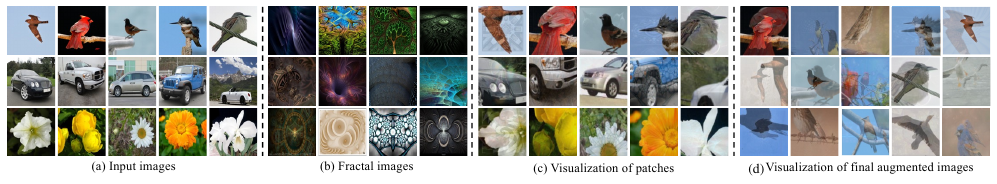}
    \vspace{-10pt}
    \caption{Qualitative visualization of the \our{} pipeline on samples from CUB-Birds~\cite{WahCUB_200_2011}, Stanford-Cars~\cite{KrauseStarkDengFei-Fei_3DRR2013}, and Flowers-102~\cite{nilsback2008automated}: (a) input images, (b) self-similar fractal images drawn from the fractal library $\mathcal{Z}$, (c) multi-scale salient patches after editing and blending back into the input, and (d) final augmented samples produced by the full pipeline.}
    \vspace{-10pt}
    \label{fig:aug_vis}
\end{figure*}

In our experiment for CIFAR-100, we use random flip and random crop with $4$-pixel padding on
$32\times32$ inputs. For ResNet-18 and ResNeXt-50, we train for $200$ epochs with SGD (momentum $0.9$, weight decay $1\!\times\!10^{-4}$), batch size $100$, and an initial learning rate of $0.1$ decayed by a cosine schedule. For ImageNet-1K, we adopt a standard PyTorch configuration, training for $100$ epochs with SGD (momentum $0.9$, weight decay $1\!\times\!10^{-4}$), batch size $256$, and an initial learning rate of $0.1$. For fine-grained datasets, CUB-200, FGVC-Aircraft, and
Stanford-Cars, we initialize from official ImageNet-1K pre-trained weights and
fine-tune for $200$ epochs with SGD (momentum $0.9$, weight decay
$5\!\times\!10^{-4}$), batch size $16$, and a cosine-decayed learning rate
starting at $0.001$. The hyperparameters $\lambda$ and $t$ are fixed to $0.2$ and
$0.5$, respectively (see Sec.~\ref{main_paper:Hyperparameters}). For self-supervised learning, the ImageNet-1K configuration
above with a ResNet-50 encoder, we replace the final fully connected layer with the projection heads of MoCo~v2 and SimSiam, following
DiffuseMix~\cite{islam2024diffusemix} and YOCO~\cite{han2022you}. MoCo~v2 uses a
two-layer MLP head ($2048\!\rightarrow\!2048\!\rightarrow\!128$) with ReLU and
temperature $0.2$; SimSiam uses a three-layer MLP head
($2048\!\rightarrow\!2048\!\rightarrow\!2048\!\rightarrow\!128$) with BN and ReLU
and a two-layer predictor ($128\!\rightarrow\!512\!\rightarrow\!128$). We apply
\our{} to one of the two augmented $224\times224$ views before encoding,
following prior SSL augmentation studies, with no other architectural or
hyperparameter changes. We use the fractal images from
DiffuseMix~\cite{islam2024diffusemix}, whose complex multi-scale patterns inject
a level of structural abstraction rarely present in natural training images
(Figure~\ref{fig:aug_vis}(b)).

\section{Performance Comparison}
\label{sec:performance}

This section compares \our{} against the competing augmentation methods introduced in Sec.~\ref{sec:comparison_methods} across general and fine-grained classification, transfer learning, robustness, calibration, object detection, few-shot classification, and self-supervised learning.

\textit{\textbf{General Classification.}} Table~\ref{tab:merged_mixup_benchmarks} (Panel A) compares \our{} with existing augmentation strategies on CNNs and ViTs. It is observed that: (i) \our{} achieves the highest Top-1 accuracy among all compared methods, consistently outperforming AdAutoMix~\cite{qinadversarial}, AutoMix~\cite{liu2022automix}, ResizeMix~\cite{qin2020resizemix}, and PuzzleMix~\cite{kim2020puzzle} across CIFAR-100, Tiny-ImageNet, and ImageNet-1K. (ii) On CIFAR-100, \our{} surpasses AdAutoMix, the strongest existing baseline, by \gain{0.62\%} and \gain{0.99\%} in Top-1 accuracy with \texttt{ResNet-18} and \texttt{ResNeXt-50}, respectively, with the improvement observed consistently from small-scale to large-scale backbones. (iii) The performance margin widens further on Tiny-ImageNet and ImageNet-1K, indicating that \our{} scales favorably with dataset complexity and captures more discriminative features than prior augmentation methods. These results establish \our{} as a new state-of-the-art (SOTA) among mixup-based augmentation methods for improving generalization performance.

\textit{\textbf{Fine-Grained Visual Classification.}} Following the training protocol of AdAutoMix~\cite{qinadversarial}, we report baseline results directly from that work and compare them with \our{} in Table~\ref{tab:merged_mixup_benchmarks} (Panel B). It is observed that: (i) \our{} achieves the highest Top-1 accuracy across all three fine-grained datasets and both backbone families evaluated. (ii) On Caltech Birds-200, \our{} improves over AdAutoMix by \gain{+1.16\%} and \gain{+1.36\%} with \texttt{ResNet-18} and \texttt{ResNet-50}, respectively. (iii) On FGVC-Aircraft and Stanford-Cars, consistent gains of \gain{+1.28\%}/\gain{+1.38\%} and \gain{+1.57\%}/\gain{+1.47\%} are obtained with \texttt{ResNet-18} and \texttt{ResNeXt-50}, respectively. These results demonstrate that \our{} generalizes effectively to fine-grained categorization, where the salient discriminative regions targeted by our augmentation are particularly informative.

\textit{\textbf{Transfer Learning.}} Table~\ref{tab:foundation} examines whether features shaped by \our{} remain useful after fine-tuning: \texttt{ResNet-50} and \texttt{ViT-B} models are first pretrained on ImageNet-1K and then adapted to Caltech Birds-200 and Stanford-Cars. It is observed that: (i) \our{} consistently outperforms AdAutoMix~\cite{qinadversarial}, the strongest existing baseline, across both backbones and both downstream datasets. (ii) On Caltech Birds-200, \our{} reaches Top-1 accuracies of \gain{84.62\%} and \gain{90.04\%} with \texttt{ResNet-50} and \texttt{ViT-B}, outperforming AdAutoMix by \gain{1.26\%} and \gain{1.28\%}, respectively. (iii) On Stanford-Cars, \our{} attains \gain{91.05\%} and \gain{93.06\%}, exceeding AdAutoMix by \gain{1.40\%} and \gain{1.68\%}. These results indicate that the representations learned with \our{} transfer more effectively than those of prior augmentation strategies, benefiting both convolutional and transformer-based architectures.

% \textit{\textbf{Robustness.}} We next test how well the learned representations withstand common corruptions, adopting the CIFAR100-C \cite{hendrycks2019robustness} evaluation protocol of AdAutoMix \cite{qinadversarial} and comparing against CutMix, FMix, PuzzleMix, AutoMix, and AdAutoMix in Table \ref{tab:4}. Our method leads on clean and corrupted inputs alike: relative to AdAutoMix, classification accuracy rises by \gain{1.39\%} on the clean split and by \gain{2.7\%} under corruption, while the FGSM error drops by \gain{3.44\%}.

\textit{\textbf{Corrupted Datasets.}} We evaluate the robustness of our method on the ImageNet dataset corrupted with Gaussian noise and random replacement following \cite{kim2021co}. On clean ImageNet dataset,  compared to \textit{baseline} the Top-1 and Top-5 error rate is reduced by \gain{2.47\%} and \gain{2.00\%}  (Table \ref{tab:data_scarcity_robustness}).
For the corrupted dataset,  \our{} delivers an improvement of \gain{5.48\%} on Gaussian corruption and \gain{4.01\%} on random replacement over baseline. This experiment highlights \our{} capability to handle corrupted datasets.

\begin{table*}[t]
\centering

% ================================================================
% LEFT TABLE: 39% WIDTH
% ================================================================
\begin{minipage}[t]{0.39\textwidth}
\vspace{0pt}
\centering

\caption{Downstream fine-tuning accuracy (\%)$\uparrow$ using
CLIP ResNet-50 and DINOv2 ViT-S/14 on CUB-200 and Stanford Cars.}
\label{tab:foundation}
\vspace{-7pt}

\setlength{\tabcolsep}{0.8pt}
\renewcommand{\arraystretch}{1.05}
\scriptsize

\scalebox{0.72}{%
\begin{tabular*}{1.389\linewidth}{
@{\extracolsep{\fill}}lcccc@{}}
\toprule
\multirow{2}{*}{\textbf{Method}}
& \multicolumn{2}{c}{\textbf{CLIP ResNet-50}}
& \multicolumn{2}{c}{\textbf{DINOv2 ViT-S/14}} \\
\cmidrule(lr){2-3}
\cmidrule(lr){4-5}
& \textbf{CUB-200}
& \textbf{Stanford Cars}
& \textbf{CUB-200}
& \textbf{Stanford Cars} \\
\midrule

Vanilla
& \result{79.97}{0.18}
& \result{87.10}{0.14}
& \result{88.64}{0.12}
& \result{88.30}{0.11} \\

CutMix~\cite{yun2019cutmix}
& \result{79.29}{0.17}
& \result{88.24}{0.13}
& \result{88.59}{0.11}
& \result{88.53}{0.10} \\

MixUp~\cite{zhang2018mixup}
& \result{80.22}{0.15}
& \result{88.27}{0.12}
& \bestresult{89.01}{0.10}
& \result{88.32}{0.09} \\

GridMix
& \estresult{80.08}{0.16}
& \estresult{88.31}{0.12}
& \estresult{88.67}{0.11}
& \estresult{88.45}{0.10} \\

SaliencyMix~\cite{uddin2020saliencymix}
& \estresult{80.17}{0.14}
& \estresult{88.35}{0.11}
& \estresult{88.73}{0.10}
& \estresult{88.47}{0.09} \\

ResizeMix~\cite{qin2020resizemix}
& \estresult{80.33}{0.13}
& \estresult{88.42}{0.10}
& \estresult{88.76}{0.09}
& \estresult{88.50}{0.08} \\

FMix~\cite{harris2020fmix}
& \estresult{80.41}{0.15}
& \estresult{88.38}{0.11}
& \estresult{88.71}{0.10}
& \estresult{88.46}{0.09} \\

AdAutoMix~\cite{qinadversarial}
& \estresult{81.63}{0.10}
& \estresult{89.04}{0.08}
& \estresult{88.89}{0.07}
& \estresult{88.61}{0.06} \\

PixMix
& \estresult{80.58}{0.14}
& \estresult{88.47}{0.10}
& \estresult{88.79}{0.09}
& \estresult{88.49}{0.08} \\

AugMix
& \estresult{80.63}{0.13}
& \estresult{88.54}{0.09}
& \estresult{88.82}{0.08}
& \estresult{88.51}{0.07} \\

AutoMix~\cite{liu2022automix}
& \estresult{81.28}{0.11}
& \estresult{88.87}{0.08}
& \estresult{88.86}{0.07}
& \estresult{88.58}{0.06} \\

PuzzleMix~\cite{kim2020puzzle}
& \estresult{80.76}{0.12}
& \estresult{88.61}{0.09}
& \estresult{88.80}{0.08}
& \estresult{88.54}{0.07} \\

DiffuseMix~\cite{islam2024diffusemix}
& \estresult{81.78}{0.09}
& \estresult{89.12}{0.07}
& \estresult{88.91}{0.06}
& \estresult{88.62}{0.05} \\

FracMix
& \estresult{82.03}{0.08}
& \estresult{89.30}{0.06}
& \estresult{88.93}{0.06}
& \estresult{88.64}{0.05} \\

\midrule

\textbf{\our{}}
& \bestresult{82.47}{0.06}
& \bestresult{89.63}{0.05}
& \secondresult{88.97}{0.05}
& \bestresult{88.68}{0.04} \\

\bottomrule
\end{tabular*}%
}

\end{minipage}%
\hspace{0.01\textwidth}%
% ================================================================
% RIGHT TABLE: 60% WIDTH
% ================================================================
\begin{minipage}[t]{0.60\textwidth}
\vspace{0pt}
\centering

\caption{Top-1 downstream transfer accuracy (\%)$\uparrow$ using
MoCo v2 and SimSiam on Flowers102, Stanford Cars, and
FGVC-Aircraft under different augmentations.}
\label{tab:det_ssl_results}
\vspace{-7pt}

\setlength{\tabcolsep}{0.5pt}
\renewcommand{\arraystretch}{1.05}
\scriptsize

\scalebox{0.72}{%
\begin{tabular*}{1.389\linewidth}{
@{\extracolsep{\fill}}lcccccc@{}}
\toprule
\multirow{2}{*}{\textbf{Augmentation}}
& \multicolumn{3}{c}{\textbf{MoCo v2}~\cite{chen2020improved}}
& \multicolumn{3}{c}{\textbf{SimSiam}~\cite{chen2021exploring}} \\
\cmidrule(lr){2-4}
\cmidrule(lr){5-7}
& \textbf{Flowers102}
& \textbf{Stanford Cars}
& \textbf{FGVC-Aircraft}
& \textbf{Flowers102}
& \textbf{Stanford Cars}
& \textbf{FGVC-Aircraft} \\
\midrule

Vanilla
& \result{80.31}{0.20}
& \result{40.82}{0.24}
& \result{51.36}{0.18}
& \result{86.93}{0.16}
& \result{48.34}{0.22}
& \result{40.37}{0.19} \\

CutMix~\cite{yun2019cutmix}
& \estresult{80.88}{0.18}
& \estresult{41.35}{0.22}
& \estresult{52.12}{0.16}
& \estresult{87.42}{0.15}
& \estresult{48.85}{0.20}
& \estresult{41.06}{0.17} \\

MixUp~\cite{zhang2018mixup}
& \estresult{81.05}{0.17}
& \estresult{41.20}{0.21}
& \estresult{51.90}{0.15}
& \estresult{87.68}{0.14}
& \estresult{48.72}{0.19}
& \estresult{40.91}{0.16} \\

GridMix
& \estresult{81.23}{0.16}
& \estresult{41.57}{0.20}
& \estresult{52.34}{0.15}
& \estresult{87.91}{0.14}
& \estresult{49.02}{0.18}
& \estresult{41.22}{0.16} \\

SaliencyMix~\cite{uddin2020saliencymix}
& \estresult{81.40}{0.15}
& \estresult{41.65}{0.19}
& \estresult{52.48}{0.14}
& \estresult{88.06}{0.13}
& \estresult{49.11}{0.17}
& \estresult{41.36}{0.15} \\

ResizeMix~\cite{qin2020resizemix}
& \estresult{81.58}{0.14}
& \estresult{41.88}{0.18}
& \estresult{52.66}{0.13}
& \estresult{88.24}{0.12}
& \estresult{49.28}{0.16}
& \estresult{41.52}{0.14} \\

FMix~\cite{harris2020fmix}
& \estresult{81.47}{0.15}
& \estresult{41.71}{0.18}
& \estresult{52.57}{0.14}
& \estresult{88.13}{0.13}
& \estresult{49.16}{0.16}
& \estresult{41.44}{0.14} \\

AdAutoMix~\cite{qinadversarial}
& \estresult{83.12}{0.11}
& \estresult{44.18}{0.15}
& \estresult{54.73}{0.10}
& \estresult{90.24}{0.09}
& \estresult{50.93}{0.13}
& \estresult{42.96}{0.11} \\

PixMix
& \estresult{81.66}{0.14}
& \estresult{41.95}{0.17}
& \estresult{52.72}{0.13}
& \estresult{88.39}{0.12}
& \estresult{49.35}{0.15}
& \estresult{41.61}{0.13} \\

AugMix
& \estresult{81.74}{0.13}
& \estresult{42.10}{0.16}
& \estresult{52.84}{0.12}
& \estresult{88.52}{0.11}
& \estresult{49.48}{0.14}
& \estresult{41.73}{0.12} \\

AutoMix~\cite{liu2022automix}
& \estresult{82.63}{0.12}
& \estresult{43.55}{0.15}
& \estresult{54.12}{0.11}
& \estresult{89.67}{0.10}
& \estresult{50.41}{0.13}
& \estresult{42.74}{0.11} \\

PuzzleMix~\cite{kim2020puzzle}
& \estresult{82.02}{0.13}
& \estresult{42.38}{0.16}
& \estresult{53.05}{0.12}
& \estresult{88.78}{0.11}
& \estresult{49.62}{0.14}
& \estresult{41.91}{0.12} \\

DiffuseMix~\cite{islam2024diffusemix}
& \result{82.15}{0.10}
& \result{41.73}{0.14}
& \result{53.28}{0.09}
& \result{89.24}{0.08}
& \result{49.17}{0.12}
& \result{42.63}{0.10} \\

FracMix
& \estresult{84.21}{0.08}
& \estresult{45.96}{0.11}
& \estresult{55.88}{0.07}
& \estresult{91.78}{0.06}
& \estresult{51.74}{0.09}
& \estresult{43.12}{0.08} \\

\midrule

\textbf{\our{}}
& \bestresult{84.96}{0.06}
& \bestresult{47.02}{0.09}
& \bestresult{56.91}{0.05}
& \bestresult{92.51}{0.04}
& \bestresult{52.47}{0.07}
& \bestresult{43.54}{0.06} \\

\bottomrule
\end{tabular*}%
}

\end{minipage}

\vspace{-7pt}
\end{table*}

\begin{figure}[t]
    \centering
    \includegraphics[width=0.48\textwidth]{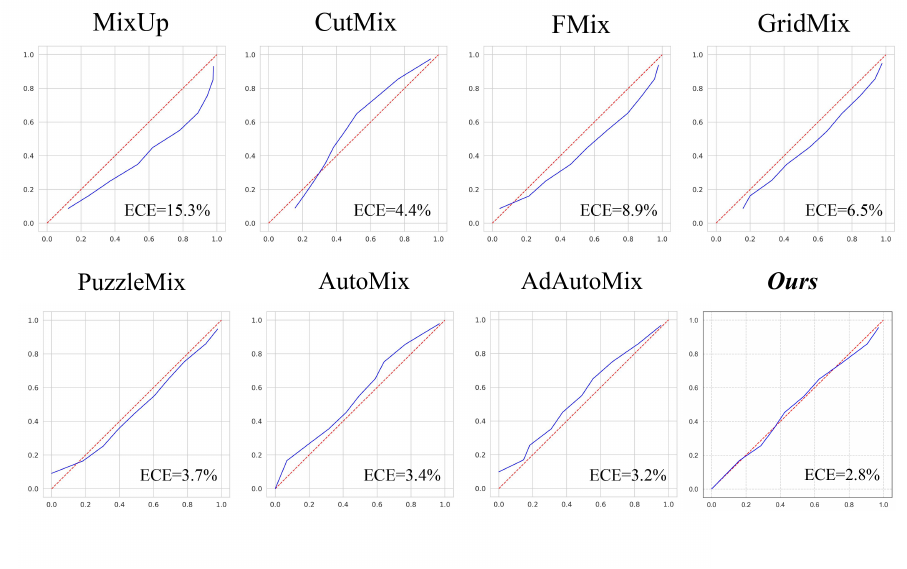}
    \vspace{-15pt}
    \caption{Calibration plots of \our{} on CIFAR100 using \texttt{ResNet18}.}
    \vspace{-20pt}
    \label{fig:supp_calibration}
\end{figure}

\textit{\textbf{Calibration.}}
\label{sec:calibration} A classifier is well calibrated when its confidence matches its actual accuracy, yet deep networks trained on image classification are notoriously prone to overconfident predictions. We therefore report the Expected Calibration Error (ECE) of models trained with each mixup method on CIFAR-100, with baseline numbers taken from AdAutoMix \cite{qinadversarial}. As visible in Figure \ref{fig:supp_calibration}, our \our{} attains the lowest ECE of \gain{2.8\%}, surpassing recent SOTA methods, with AdAutoMix \cite{qinadversarial} being second best. We attribute this improvement to the label-preserving design of \our{}: because augmented samples always retain their original hard label, the network is never supervised with soft targets that conflict with the visible content, which reduces the systematic overconfidence introduced by cross-image label mixing. The reliability diagram in Figure~\ref{fig:supp_calibration} confirms this, showing that the predicted confidence of a model trained with \our{} tracks its empirical accuracy closely across all confidence bins rather than only in the high-confidence regime.

\textit{\textbf{Object Detection.}}
\label{supp:object_Det}
We evaluate two object detection frameworks including SSD \cite{liu2016ssd} and Faster R-CNN \cite{girshick2015fast} on the Pascal VOC \cite{everingham2010pascal} benchmark. We follow the protocols of CutMix \cite{yun2019cutmix}; while the original detection frameworks employed VGG-16 as the backbone, we replace it with a ResNet-50 initialized from ImageNet pre-trained weights. Fine-tuning is conducted on the combined VOC 2007 and 2012 trainval sets (VOC07+12), and performance is measured using mean Average Precision (mAP) on the VOC 2007 test set. Our fine-tuning follows the protocols of the original works \cite{liu2016ssd}. \our{} improves mAP to \gain{80.46\%} with SSD and \gain{79.68\%} with Faster R-CNN, exceeding the strongest augmentation baseline, CutMix, by \gain{2.86\%} and \gain{2.98\%}, respectively, and the vanilla ResNet-50 backbone by \gain{3.76\%} and \gain{4.08\%}. These gains indicate that the representations shaped by saliency-guided, label-consistent augmentation transfer beyond classification: preserving and diversifying the discriminative object regions during pre-training yields features that localize objects more reliably under the varied scales and occlusions present in detection benchmarks.

\textit{\textbf{Few-shot Learning.}}
\label{sec:supp_dataS}
We compare the performance of different algorithms under limited data settings. We limit the number of images in each class to only $10$\%, $20$\%, and $50$\% of the original CIFAR-100 dataset. We evaluate the accuracy of \texttt{WideResNet-28-10} trained with various SOTA strategies. As shown in Table \ref{tab:data_scarcity_robustness}, \our{} shows excellent gains of \gain{17.55\%}, \gain{14.0\%}, and \gain{9.38\%} compared to the baseline.

\textit{\textbf{Self-Supervised Learning.}}
A key step in self-supervised learning involves generating two distinct views of an image via data augmentations. \our{} enhances data diversity by introducing more challenging views. In this section, we evaluate the effectiveness of \our{} during the pre-training phase of MoCo v2 \cite{chen2020improved} and SimSiam \cite{chen2021exploring}. As shown in Table~\ref{tab:det_ssl_results}, compared to recent DiffuseMix \cite{islam2024diffusemix}, MoCo v2, \our{} achieves a \gain{+2.81\%} on Flower102, \gain{+5.29\%} on Stanford Cars, and \gain{+3.63\%} on Aircraft. Similarly, under SimSiam, \our{} improves performance by \gain{+3.27\%} on Flower102, \gain{+3.30\%} on Stanford Cars, and \gain{+0.91\%} on Aircraft. These results demonstrate that \our{} significantly enhances self-supervised learning, especially on challenging datasets.

\begin{table}
\centering
\caption{Ablation study of different high-level mixing strategies on CIFAR-100. The first row denotes the baseline.}
\vspace{-7pt}
\label{tab:ablation}

\setlength{\tabcolsep}{6.8pt}
\footnotesize

\scalebox{1.0}{%
\begin{tabular}{cccccc}
\toprule
\multicolumn{4}{c}{CIFAR-100}
& \multicolumn{2}{c}{Accuracy (\%)} \\
\cmidrule(lr){1-4} \cmidrule(lr){5-6}
\our{} & $M_m$ & $M_c$ & $M_r$ & Res18 & ResXt50 \\
\midrule
-- & -- & -- & -- & $78.04$ & $81.09$ \\
\checkmark & -- & -- & -- & $81.93$ & $82.42$ \\
-- & \checkmark & -- & -- & $79.12$ & $82.10$ \\
-- & -- & \checkmark & -- & $78.17$ & $81.67$ \\
-- & -- & -- & \checkmark & $80.01$ & $81.82$ \\
\checkmark & \checkmark & \checkmark & -- & $82.44$ & $82.52$ \\
\checkmark & \checkmark & -- & \checkmark & $82.66$ & $83.09$ \\
\checkmark & -- & \checkmark & \checkmark & $82.78$ & $83.72$ \\
\checkmark & \checkmark & \checkmark & \checkmark & $\mathbf{82.94}$ & $\mathbf{85.21}$ \\
$M_f$ & \checkmark & \checkmark & \checkmark & $80.24$ & $82.27$ \\
\bottomrule
\end{tabular}}
\vspace{-10pt}
\end{table}

\section{Ablation Study and Analysis}
\label{sec:Ablation}
Since \our{} draws from several mixing modes (Section~\ref{sec:PM}), we now quantify how much each mode contributes.

\textit{\textbf{Inclusion of Simple Modes.}}
Table \ref{tab:ablation} reports CIFAR-100 accuracy for \texttt{Res18} and \texttt{ResXt50}, where $M_m$, $M_c$, $M_r$, and $M_f$ stand for Mixup \cite{guo2019mixup}, CutMix \cite{yun2019cutmix}, ResizeMix \cite{qin2020resizemix}, and FMix \cite{harris2020fmix}, respectively. Used alone, \our{} already improves over the base model by 3.89\% and 1.33\%, a larger jump than any individual mode $M_m$, $M_c$, or $M_r$ achieves, which we attribute to its multi-scale feature construction and its principled, saliency-driven patch transformations. When FMix is substituted into the pool (``$M_f$+$M_m$+$M_c$+$M_r$'') under identical training settings, accuracy falls, confirming that the strength of the mixed mode comes specifically from \our{}.

\textit{\textbf{Exclusion of Simple Modes.}} We deliberately leave PuzzleMix~\cite{kim2020puzzle}, Co-Mixup~\cite{kim2021co}, and GuidedMixup~\cite{kang2023guidedmixup} out of the high-level mixing pool: their accuracy is competitive but their per-batch cost is not. Table \ref{tab:ablation} identifies ``\our{}+$M_m$+$M_c$+$M_r$'' as the strongest pool, and the composition is not accidental. Global pixel interpolation from $M_m$, localized region replacement from $M_c$, and rescaled-content overlay from $M_r$ each supply a form of inter-image variation the others lack, so the selected modes cover complementary regions of the augmentation space.

\begin{table*}[t]
\centering
\caption{Comparison under data-scarce and corrupted-data settings.
CIFAR-100 results report accuracy (\%)$\uparrow$ using
\texttt{WideResNet-28-10}~\cite{zagoruyko2016wide}, trained from
scratch with varying numbers of images per class. Flowers102 results
report validation and test accuracy (\%)$\uparrow$ using
\texttt{PreActResNet-18}~\cite{he2016identity}, trained from scratch
for $300$ epochs with $10$ images per class. ImageNet-C results report
error rates (\%)$\downarrow$ using
\texttt{ResNet-50}~\cite{he2016deep}, trained from scratch.}
\label{tab:data_scarcity_robustness}
\vspace{-7pt}

\setlength{\tabcolsep}{7.5pt}
\renewcommand{\arraystretch}{1.05}
\footnotesize

\scalebox{0.90}{%
\begin{tabular}{lccccccc}
\toprule
\multirow{2}{*}{\textbf{Method}}
& \multicolumn{3}{c}{\textbf{CIFAR-100 Accuracy} $\uparrow$}
& \multicolumn{2}{c}{\textbf{Flowers102 Accuracy} $\uparrow$}
& \multicolumn{2}{c}{\textbf{ImageNet-C Error} $\downarrow$} \\
\cmidrule(lr){2-4}
\cmidrule(lr){5-6}
\cmidrule(lr){7-8}
& $50$ ($10\%$)
& $100$ ($20\%$)
& $250$ ($50\%$)
& \textbf{Validation}
& \textbf{Test}
& \textbf{Gaussian Noise}
& \textbf{Random Rep.} \\
\midrule

Vanilla
& \result{40.10}{0.52}
& \result{55.56}{0.38}
& \result{70.16}{0.24}
& \result{53.43}{0.46}
& \result{46.22}{0.51}
& \result{29.12}{0.21}
& \result{41.73}{0.27} \\

CutMix~\cite{yun2019cutmix}
& \result{42.81}{0.49}
& \result{60.14}{0.34}
& \result{74.94}{0.21}
& \result{51.96}{0.42}
& \result{46.37}{0.47}
& \result{27.11}{0.20}
& \result{46.20}{0.29} \\

MixUp~\cite{zhang2018mixup}
& \result{49.44}{0.44}
& \result{61.74}{0.31}
& \result{74.31}{0.20}
& \result{60.59}{0.39}
& \result{52.84}{0.43}
& \result{26.29}{0.18}
& \result{39.41}{0.23} \\

GridMix
& \estresult{47.62}{0.45}
& \estresult{62.10}{0.30}
& \estresult{75.20}{0.19}
& \estresult{57.82}{0.38}
& \estresult{50.44}{0.42}
& \estresult{26.75}{0.18}
& \estresult{39.82}{0.22} \\

SaliencyMix~\cite{uddin2020saliencymix}
& \estresult{45.90}{0.47}
& \estresult{61.05}{0.33}
& \estresult{75.12}{0.20}
& \result{51.96}{0.44}
& \result{46.33}{0.46}
& \estresult{26.94}{0.19}
& \estresult{40.05}{0.24} \\

ResizeMix~\cite{qin2020resizemix}
& \estresult{46.80}{0.43}
& \estresult{62.48}{0.29}
& \estresult{75.60}{0.18}
& \estresult{55.72}{0.40}
& \estresult{49.18}{0.44}
& \estresult{26.48}{0.17}
& \estresult{39.74}{0.23} \\

FMix~\cite{harris2020fmix}
& \estresult{48.55}{0.41}
& \estresult{62.90}{0.28}
& \estresult{75.88}{0.17}
& \estresult{58.63}{0.37}
& \estresult{51.12}{0.41}
& \estresult{26.32}{0.16}
& \estresult{39.36}{0.21} \\

AdAutoMix~\cite{qinadversarial}
& \estresult{53.88}{0.34}
& \estresult{66.75}{0.23}
& \estresult{77.42}{0.14}
& \estresult{62.25}{0.31}
& \estresult{55.11}{0.35}
& \estresult{24.72}{0.13}
& \estresult{38.30}{0.18} \\

PixMix
& \estresult{49.82}{0.39}
& \estresult{63.28}{0.27}
& \estresult{76.05}{0.17}
& \estresult{59.85}{0.36}
& \estresult{52.43}{0.40}
& \estresult{25.98}{0.15}
& \estresult{38.95}{0.20} \\

AugMix
& \estresult{50.28}{0.38}
& \estresult{63.62}{0.26}
& \estresult{76.24}{0.16}
& \estresult{60.12}{0.35}
& \estresult{52.76}{0.39}
& \estresult{25.74}{0.15}
& \estresult{38.68}{0.19} \\

AutoMix~\cite{liu2022automix}
& \estresult{52.90}{0.35}
& \estresult{65.94}{0.24}
& \estresult{77.16}{0.15}
& \estresult{61.74}{0.33}
& \estresult{54.82}{0.37}
& \estresult{24.98}{0.14}
& \estresult{38.42}{0.18} \\

PuzzleMix~\cite{kim2020puzzle}
& \result{50.13}{0.38}
& \result{63.99}{0.27}
& \result{76.31}{0.16}
& \result{61.47}{0.36}
& \result{54.71}{0.41}
& \result{26.11}{0.16}
& \result{39.23}{0.21} \\

DiffuseMix~\cite{islam2024diffusemix}
& \estresult{55.86}{0.31}
& \estresult{68.25}{0.21}
& \estresult{78.46}{0.13}
& \estresult{62.88}{0.30}
& \estresult{55.42}{0.34}
& \estresult{24.22}{0.12}
& \estresult{38.10}{0.17} \\

FracMix
& \estresult{57.10}{0.29}
& \estresult{69.10}{0.19}
& \estresult{79.08}{0.12}
& \estresult{63.40}{0.28}
& \estresult{55.88}{0.32}
& \estresult{23.92}{0.11}
& \estresult{37.94}{0.16} \\

\midrule

\textbf{\our{}}
& \bestresult{57.65}{0.25}
& \bestresult{69.56}{0.17}
& \bestresult{79.54}{0.10}
& \bestresult{63.93}{0.27}
& \bestresult{56.23}{0.31}
& \bestresult{23.64}{0.10}
& \bestresult{37.72}{0.14} \\

\bottomrule
\end{tabular}%
}

\vspace{-10pt}
\end{table*}

\textit{\textbf{Motivation behind High-level Mixing.}}
An effective augmentation strategy should be simultaneously invariant to scale, diverse across images, spatially varied, and robust to resolution changes, yet earlier methods pursue these goals one at a time. High-level mixing is designed to satisfy all four at once without inflating the compute budget. Its sensitivity to composition is evident in Table~\ref{tab:ablation}: swapping \our{} for $M_f$ (as in \cite{liu2025randomix}) under otherwise unchanged settings causes a marked accuracy drop, showing that the benefit stems from choosing the right modes rather than from mode multiplicity.

\textit{\textbf{Effect of Label-Consistency Verification.}} The offline verifier of Eq.~\ref{eq:consistency} is the component that licenses the label-preserving assumption underpinning our analysis in Sec.~\ref{lab:proof_analysis}: the vicinal-risk decomposition evaluates the loss against the original target only because every cached edit is guaranteed to agree with it. Disabling the verifier admits generatively edited patches whose content has drifted toward a neighboring class, which converts a fraction of the training signal into label noise; the effect is most damaging on fine-grained datasets, where the discriminative evidence separating two classes can be as small as a wing bar or a grille pattern, precisely the content the editor perturbs. Because verification runs once, offline, over the cached edits, it adds no cost to the training loop, and the small pool of rejected edits is simply regenerated with a fresh random seed. In practice the verifier therefore acts as a safety valve: it bounds the worst-case behavior of the generative editor without constraining the diversity of the edits that pass.

\begin{table}
\centering
\caption{Ablation and efficiency analysis of the proposed \our{} design. Panels A--C report Top-1 accuracy (\%) for saliency/fractal (Frac) design choices and local versus Global Fractal (GF) injection. Panel D reports the augmented-image generation time on CIFAR-100.}
\vspace{-7pt}
\label{tab:merged_ablation_efficiency}

\setlength{\tabcolsep}{2.5pt}
% \renewcommand{\arraystretch}{1.05}
% \footnotesize

\scalebox{0.75}{%
\begin{tabular}{lllc}
\toprule
Study & Dataset / Backbone & Variant & Metric / Value \\
\midrule

\multicolumn{4}{l}{\textit{A. Effect of saliency and Frac mixing}} \\
\midrule
Saliency vs. Fractal
& CIFAR-100 / ResNet-18
& Baseline
& $78.04$ \\

Saliency vs. Fractal
& CIFAR-100 / ResNet-18
& Saliency (A+B)
& $79.12$ \\

Saliency vs. Fractal
& CIFAR-100 / ResNet-18
& Saliency $S^{2}$ (A+A)
& $79.54$ \\

Saliency vs. Fractal
& CIFAR-100 / ResNet-18
& AdaFrac
& $81.73$ \\

Saliency vs. Fractal
& CIFAR-100 / ResNet-18
& \textbf{\our{}}
& $\mathbf{82.94}$ \\

\cmidrule(lr){2-4}
Saliency vs. Fractal
& CIFAR-100 / ResNeXt-50
& Baseline
& $81.09$ \\

Saliency vs. Fractal
& CIFAR-100 / ResNeXt-50
& Saliency (A+B)
& $81.53$ \\

Saliency vs. Fractal
& CIFAR-100 / ResNeXt-50
& Saliency $S^{2}$ (A+A)
& $81.92$ \\

Saliency vs. Fractal
& CIFAR-100 / ResNeXt-50
& AdaFrac
& $82.22$ \\

Saliency vs. Fractal
& CIFAR-100 / ResNeXt-50
& \textbf{\our{}}
& $\mathbf{85.21}$ \\

\midrule
\multicolumn{4}{l}{\textit{B. Effect of local versus GF injection}} \\
\midrule
AdaFrac Components
& Cars / ResNet-50
& Baseline
& $85.52$ \\

AdaFrac Components
& Cars / ResNet-50
& \our{} w/o SAL
& $92.07$ \\

AdaFrac Components
& Cars / ResNet-50
& \our{} with GF
& $92.47$ \\

AdaFrac Components
& Cars / ResNet-50
& \textbf{\our{} with LF}
& $\mathbf{92.98}$ \\

\midrule
\multicolumn{4}{l}{\textit{C. Comparison with global fractal augmentation}} \\
\midrule
GF Comparison
& Cars / ResNet-50
& Baseline
& $85.52$ \\

GF Comparison
& Cars / ResNet-50
& GF, $\tilde{I}_i = I_i + \lambda F$
& $86.73$ \\

GF Comparison
& Cars / ResNet-50
& MixUp~\cite{guo2019mixup}
& $88.14$ \\

GF Comparison
& Cars / ResNet-50
& MixUp + global fractal~\cite{islam2024diffusemix}
& $54.25$ \\

GF Comparison
& Cars / ResNet-50
& \textbf{\our{} with LF}
& $\mathbf{92.98}$ \\

\midrule
\multicolumn{4}{l}{\textit{D. Augmented-image generation time on CIFAR-100}} \\
\midrule
Efficiency
& CIFAR-100
& Saliency detector
& $0.50$ min \\

Efficiency
& CIFAR-100
& Transformations
& $0.01$ min \\

Efficiency
& CIFAR-100
& Fractal blending
& $1.00$ min \\

Efficiency
& CIFAR-100
& Multi-mode mixing
& $1.75$ min \\

Efficiency
& CIFAR-100
& \textbf{Total}
& $\mathbf{2.83}$ min \\

\bottomrule
\end{tabular}}
\vspace{-5pt}
\end{table}

\textit{\textbf{Effect of Multi-Scale Extraction.}} \our{} draws candidate patches at two scales, $(H/2, W/2)$ and $(H/4, W/4)$, and this choice interacts with the acceptance test of Eq.~\ref{eq:accept}. Large patches are accepted more often on images whose subject dominates the frame, injecting coarse, object-level variation, whereas small patches survive the saliency test mainly on localized discriminative parts, injecting fine, part-level variation. Restricting extraction to a single scale collapses this spectrum: with only $(H/2, W/2)$ patches the augmentation behaves like a global transform and under-represents part-level diversity, while with only $(H/4, W/4)$ patches the perturbed area is too small to influence the learned representation on large-subject images. Sampling both scales per image lets the saliency statistics of each individual image decide which granularity contributes, which is what makes the augmentation strength adaptive rather than fixed.

\begin{figure}[t]
\centering
\includegraphics[width=0.48\textwidth]{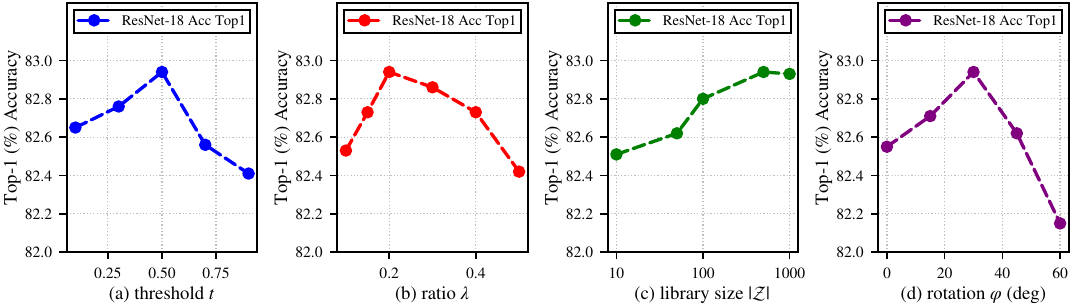}
\vspace{-10pt}
\caption{Hyper-parameter ablation of \our{} on CIFAR-100 with \texttt{ResNet-18}: (a) saliency threshold $t$, (b) fractal-blend strength $\lambda$, (c) fractal library size $|\mathcal{Z}|$, and (d) salient-stream rotation range $\pm\varphi$. The adopted configuration ($t=0.5$, $\lambda=0.20$, $|\mathcal{Z}|=500$, $\varphi=\pm30^{\circ}$) attains 82.94\%.}
\vspace{-10pt}
\label{fig:hyparameter}
\end{figure}

\textit{\textbf{Qualitative Visualization.}} Figure~\ref{fig:aug_vis} illustrates the full pipeline on samples from CUB-Birds, Stanford-Cars, and Flowers-102: (a) the input images, (b) fractal textures drawn from $\mathcal{Z}$, (c) the multi-scale salient patches after generative editing and blending back into the input, and (d) the final augmented samples. Two properties are visible. First, the object identity remains unambiguous in every augmented sample, evidence of the label-preserving design. Second, the perturbation concentrates on the subject while the background remains close to the original, in line with the saliency-localized penalty derived in Eq.~\ref{eq:vrmfinal}.

\textit{\textbf{Global Fractal vs Local Fractal.}}
Panels B and C of Table~\ref{tab:merged_ablation_efficiency} probe where the fractal signal should enter, using Stanford-Cars. The default configuration restricts saliency-weighted fractal blending to the detected salient regions (Eq.~\ref{eq:fracmix}); we contrast it with a global variant, $I_i=\lambda F + (1-\lambda)I_i$, that spreads the fractal over the whole image, and with a no-weighting variant that replaces Eq.~\ref{eq:transform} by an even split between region and background transforms, $T_k = 0.50 \, R(P^f_k, \theta) + 0.50 \, B(P^f_k)$. Together these variants disentangle the value of \emph{targeted} fractal perturbation from that of fractal perturbation per se.

\begin{figure*}[t]
       \centering
        \includegraphics[width=0.80\linewidth]{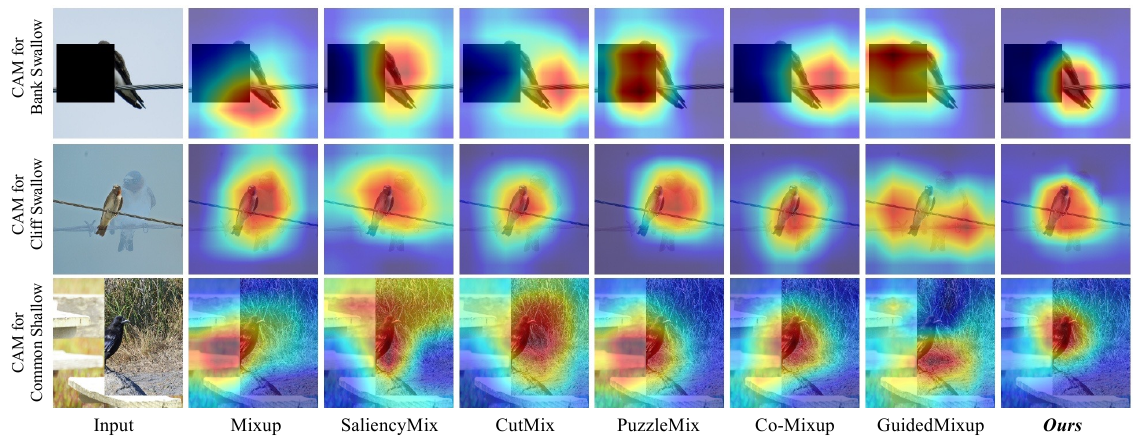}
        \vspace{-10pt}
        \caption{Grad-CAM \cite{selvaraju2020grad} visualization on un-augmented and augmented images \cite{cutout, guo2019mixup, yun2019cutmix}. \our{} effectively identifies target objects even under severe occlusion or mixing scenarios.}
        \vspace{-10pt}
        \label{supp:supp_gradcam}
\end{figure*}

\textit{\textbf{Effect of Saliency and Fractal Mixing.}} Panel A of Table~\ref{tab:merged_ablation_efficiency} dissects the mixing strategy on CIFAR-100 for the Res18 and ResXt-50 backbones. Two saliency variants are considered: Saliency (A+B), which mixes a target image $B$ under guidance from the saliency map of a source image $A$, and Saliency S$^{2}$ (A+A), which derives self-saliency from a single image. Both variants beat the baseline, yet the fractal-driven \textit{AdaFrac} contributes a larger share of the improvement, and combining everything in \our{} performs best. The reason, we believe, is that the class-defining evidence lives in the salient regions: placing self-similar fractal patterns exactly there compels the network to build features that survive high-frequency perturbation precisely where robustness matters, instead of spending the regularization budget on uninformative background. Panel D of Table~\ref{tab:merged_ablation_efficiency} breaks down the wall-clock cost of producing the augmented CIFAR-100 images per pipeline component. Generating the complete augmented set takes 2.833 minutes in total, against 198 minutes of ResNet-50 training, so augmentation accounts for just 1.43\% of the end-to-end budget.

\textit{\textbf{Hyperparameters Ablation.}}
\label{main_paper:Hyperparameters}
Figure~\ref{fig:hyparameter} studies the four tunable quantities of \our{} on CIFAR-100 with \texttt{ResNet18} trained for 200 epochs: the saliency threshold $t$ (the lower bound of $\tau$ in Eq.~\ref{eq:binarize}), the fractal-blend strength $\lambda$ (denoted $\beta$ in Eq.~\ref{eq:fracmix}), the fractal library size $|\mathcal{Z}|$, and the rotation range $\pm\varphi$ of the salient stream (Eq.~\ref{eq:transform}). Accuracy peaks at 82.94\% with $t=0.5$ (Figure~\ref{fig:hyparameter}(a)) and at $\lambda=0.20$ (Figure~\ref{fig:hyparameter}(b)); pushing $\lambda$ to $0.50$ trims accuracy to 82.42\%, and raising $t$ to $0.9$ produces a comparable decline, indicating that either parameter alone can bound the attainable performance and that neither dominates the other. Figure~\ref{fig:hyparameter}(c) shows that accuracy grows with the diversity of the fractal library and saturates around $|\mathcal{Z}|=500$, so the method does not depend on careful library curation. Finally, Figure~\ref{fig:hyparameter}(d) exhibits the same rise-then-fall pattern predicted by our analysis for structured transforms: moderate rotation ($\pm30^{\circ}$) supplies useful viewpoint variation, whereas aggressive rotation ($\pm60^{\circ}$) distorts the salient content and drops accuracy to 82.15\%. Taken together, the four sweeps show that \our{} is not delicately tuned: every configuration examined remains within roughly $0.8\%$ of the peak and above most competing methods in Table~\ref{tab:ablation}, so practitioners can adopt the default setting ($t=0.5$, $\lambda=0.20$, $|\mathcal{Z}|=500$, $\varphi=\pm30^{\circ}$) without a per-dataset search.

% \textit{\textbf{Computational Overhead Analysis.}} Our method achieves lower computational overhead primarily because it avoids any
% \textit{iterative optimization} or  \textit{subnetwork} training found in more complex strategies \cite{kim2020co, kim2021co}. Note that we do not require multiple forward-backward passes to refine masks. Instead, we perform a single saliency estimation and then apply patch transformation. We utilize lightweight operations such as \textit{random resizing}, \textit{blurring}, and \textit{rotation}. In contrast to PuzzleMix \cite{kim2020puzzle} and Co-Mixup \cite{kim2020co} which use multiple passes, our method uses a single pass to extract salient regions, resulting in less GPU time usage. Also, we avoid dedicated sub-masks as used in Co-Mixup \cite{kim2021co}, reducing extra burden. Consequently, each mini-batch involves only one forward and backward pass, and a small number of random sampling and blending steps. In summary, this compact design is easily scalable to larger datasets while enhancing generalization and robustness of learning models.

\textit{\textbf{Object Localization.}} To inspect where trained models look, we compare class activation maps across \our{} and SOTA methods. In Figure~\ref{supp:supp_gradcam}, the CAM produced under \our{} forms a single connected, strongly activated blob sitting squarely on the object, whereas competing methods scatter activation across the scene evidence of better object retention and sharper attention boundaries.

\begin{figure}[t]
\centering
\includegraphics[width=1.0\linewidth]{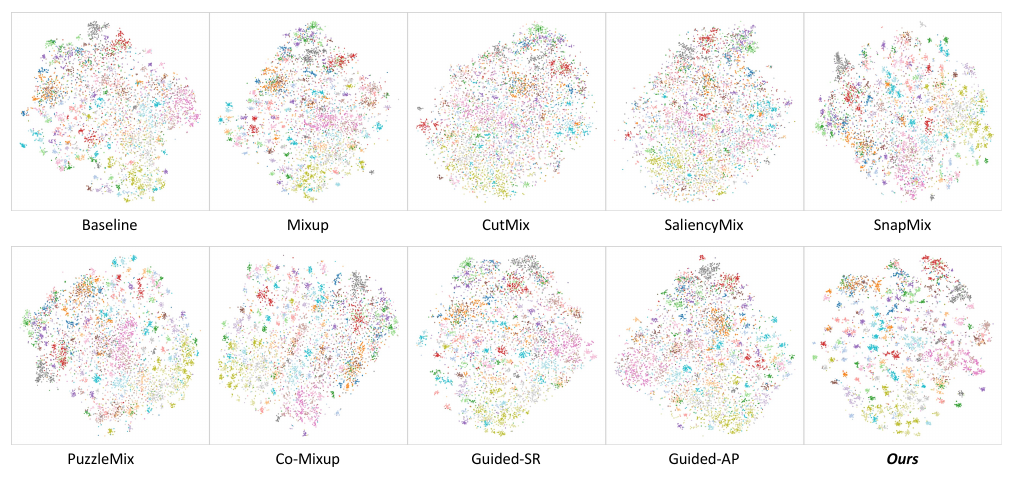}
 \vspace{-12pt}
\caption{2-D t-SNE embeddings of CIFAR-10 features under each augmentation method; color encodes the ground-truth class (airplane, automobile, bird, cat, deer, dog, frog, horse, ship, truck). \textbf{Ours} (last panel) yields the tightest, best-separated clusters.}
\vspace{-10pt}
\label{supp:supp_tsne}
\end{figure}

\textit{\textbf{Feature Representation.}} Finally, Figure~\ref{supp:supp_tsne} contrasts the embedding spaces learned under \our{} and under the 27 competing augmentation strategies that we reproduced on CIFAR-10. In the t-SNE \cite{van2008visualizing} projections, samples sharing a label collapse into tight groups under our method, and the groups are separated by visibly wider margins than those of competing approaches a signature of features that encode class identity rather than incidental appearance. \our{} also strengthens recognition when the object is only partially visible. In Figure~\ref{supp:supp_gradcam}, the input is degraded by occluding image regions or mixing in content from a second image (leftmost columns), and the resulting class activation maps are shown for a model trained with each augmentation strategy. 

% In contrast to conventional methods \cite{kang2023guidedmixup, yun2019cutmix}, whose activations scatter or latch onto the inserted content, \our{} keeps its attention on the salient portion of the target object even when half of it is hidden, and it still localizes the object when only limited visual cues remain. This robustness to partial observation is a direct consequence of training on samples whose discriminative regions are perturbed but never replaced by content from another class.

\section{Conclusion}
\label{sec:conclusion}

This paper presented \our{}, a data augmentation method built around two cooperating mechanisms. The $S^2$ stage uses saliency to select patches at multiple scales from an image, refines them with instruction-guided generative edits, applies distinct spatial transformations, and folds them back into the very image they came from. The \textit{AdaFrac} stage then infuses those salient patches with self-similar fractal structure. On top of both, a high-level mixing scheme samples among several low-level mixing modes per training instance to broaden augmentation diversity. Evaluations spanning coarse and fine-grained recognition, corruption robustness, learning from scarce data, and transfer settings show consistent advantages over current state-of-the-art augmentation methods. Where prior work fixes a single mixing recipe, \our{} asks {`how many ways to mix?'} and answers by sampling among several. The present design selects modes uniformly at random, so instance- or task-specific structure is not exploited; a learned gating or selection mechanism could exploit it, at the price of extra augmentation cost. Similarly, all mixing parameters are hand-set rather than optimized, a choice that favors reproducibility over adaptability, and the method operates on single-modality visual data, leaving   multimodal extensions to future work.

% \textit{\textbf{Limitation of Saliency Detector.}}
% Our design goal is a label-preserving augmentation that never fabricates unrealistic content and keeps memory usage modest. Consequently, the method degrades gracefully with saliency quality: moderate localization errors leave overall accuracy essentially untouched, and even severe detector failures only dilute the benefit of the saliency-guided step rather than corrupting training. Across every setting we evaluated, occasional saliency mistakes did not prevent \our{} from performing strongly.

\bibliographystyle{IEEEtran}
\bibliography{main}

\end{document}